\documentclass[journal]{IEEEtran}

\usepackage[english]{babel}
\usepackage{hyperref}
\usepackage{amsmath,epsfig}

\usepackage{tabu}
\usepackage{pdflscape}
\usepackage{dblfloatfix}
\usepackage[nameinlink,capitalize]{cleveref}
\usepackage{dblfloatfix}
\usepackage{colortbl}
\usepackage{soul}
\usepackage{placeins}
\usepackage{enumitem}
\usepackage{siunitx}
\newcommand{\bdot}{\boldsymbol\cdot}
\usepackage{amssymb}
\usepackage{pifont}
\newcommand{\xmark}{\ding{55}}
\usepackage{svg}
\author{Iris Dumeur}
\date{\today}
\title{}
\hypersetup{
 pdfauthor={Iris Dumeur},
 pdftitle={},
 pdfkeywords={},
 pdfsubject={},
 pdfcreator={Emacs 27.1 (Org mode 9.4.6)}, 
 pdflang={English}}
\begin{document}

\title{Paving the way toward foundation models for irregular and unaligned Satellite Image Time Series}
\author{Iris Dumeur, 
Silvia Valero,
Jordi Inglada
\thanks{I. Dumeur, S. Valero, J. Inglada are with Univ Toulouse 3 Paul Sabatier, Univ Toulouse, CNES/IRD/CNRS/INRAe, CESBIO, Toulouse, France
(e-mail: iris.dumeur@univ-tlse3.fr,silvia.valero-valbuena@iut-tlse3.fr, jordi.inglada@cesbio.eu).}
\thanks{This work is supported by the DeepChange project under the grant agreement ANR-DeepChange CE23}}
\maketitle
\IEEEpeerreviewmaketitle

\begin{abstract}
Although recently several foundation models for satellite remote sensing imagery have been proposed, they fail to address major challenges of operational applications. Indeed, representations that do not take into account the spectral, spatial and temporal dimensions of the data as well as the irregular or unaligned temporal sampling are of little use for most real world applications.
As a consequence, we address existing shortcomings in the design of foundation models for remote sensing data. In particular, we propose an ALIgned Sits Encoder (ALISE), a novel approach that leverages the spatial, spectral, and temporal dimensions of irregular and unaligned SITS, while producing aligned latent representations. ALISE provides easy-to-use fixed-size SITS representations which preserve the spatial resolution of the input SITS required for most mapping tasks.
Moreover, to learn informative representations of SITS, we investigate the integration of instance discrimination losses within a masked auto-encoding pre-training task, utilizing a multi-view framework. The model is pre-trained on a custom-built Sentinel-2 multi-year SITS unlabeled dataset. The genericity of the provided representations is assessed on three downstream tasks: crop segmentation, land cover segmentation, and an unsupervised crop change detection task. The results suggest that the use of ALISE's aligned representations is significantly more effective than previous SSL methods for linear probing segmentation tasks. Additionally, the experiments show the interest of using ALISE representations
for unsupervised change detection. Lastly, the impact of the pre-training hyper-parameters and the proposed method for aligning irregular and unaligned time series are examined in detail. The code, the pre-trained model as well as the datasets are released at \url{https://src.koda.cnrs.fr/iris.dumeur/alise}.
\end{abstract}
\section{Introduction}
\label{sec:org8deebc0}
Over the past decade, numerous Earth observation satellite missions have been launched to monitor the effects of climate change on land surfaces. In particular, missions combining high spectral and spatial resolutions with high temporal revisit, such as Sentinel-2 (S2) \cite{6351195}, enable an exhaustive and systematic capture of land surfaces. The acquired data correspond to Satellite Image Time Series (SITS), which are 4D objects encompassing spatial, spectral, and temporal dimensions. By nature, SITS serve as essential inputs for generating frequently large scale land cover segmentation maps required by the climate and geoscience community.
Currently, numerous Deep Learning (DL) studies have been proposed to exploit SITS for producing high-resolution maps characterizing land surfaces. For instance, several works have addressed the production of crop semantic segmentation maps \cite{garnot20_light_tempor_self_class_satel,garnot-2021-panop-segmen,pelletier-2019-tempor-convol}. Nevertheless, a major limitation of supervised DL technologies is their need for large amounts of labeled data. Existing supervised methods are therefore ill-suited to the production of large-scale maps, covering several time periods and/or the entire globe. In addition, DL methods applied to SITS are task-specific and thus not adapted to solving the multiple tasks required for Earth monitoring. 
Therefore, we propose to investigate the production of SITS representations that can be used with minimal training time and a small amount of labeled data for a wide range of Earth monitoring applications.

Building a model capable of providing multi-task representations is in line with the recent DL paradigm known as "foundation model" \cite{bommasani2022opportunitiesrisksfoundationmodels}. In remote sensing (RS), foundation models (FM) have been recently theorized as a solution to avoid the training of numerous task-specific models, exploiting the synergy of multimodal SITS, as well as mitigating the need for costly and time-consuming data annotation \cite{Zhu2024May}. To do so, the large neural network architectures of these FM need to be trained on large unlabeled datasets using self-supervised learning (SSL) strategies. FM are then supposed to have learned a generic representation relevants for numerous downstream tasks.
Training a RS FM on a large unlabeled dataset requires taking into account the specificities of SITS, which are among others \textbf{irregular} and \textbf{unaligned}. More precisely, irregularity refers to variable temporal gaps within the same time series, caused by missing acquisitions due, for example, to inappropriate atmospheric conditions (haze, fog or clouds) or sensor problems. Unalignment corresponds to different acquisition dates when comparing at least two time series. Even for using a single sensor, SITS acquired in different geographical areas are unaligned due to orbital phase shifts. 
Providing task-agnostic SITS representations, which take into account SITS specificities should also simplify the exploitation of satellite data. While recent studies have proposed SSL methodologies for SITS \cite{Tseng2023LightweightPT,dumeur-2024-self-super,astruc2024omnisat,guo2024skysense}, the existing methods do not produce:  \textbf{easy-to-use (i)}, \textbf{informative (ii)} and \textbf{generic (iii)} SITS representations. These three characteristics are further detailed below.

\textbf{(i) Providing easy-to-use representations of irregular and unaligned SITS.} FM should provide representations which serve as basis for the geosciences and climate communities. As a consequence, we consider that easy-to-use representations: are relevant without model fine-tuning, are aligned and of fixed size, and preserve the spatial resolution of the SITS. No existing SITS representation encoder \cite{yuan21_self_super_pretr_trans_satel,yuan22_sits_former,Tseng2023LightweightPT,dumeur-2024-self-super} matches those criteria. For instance,  most current methods \cite{yuan21_self_super_pretr_trans_satel,yuan22_sits_former,Tseng2023LightweightPT,guo2024skysense} provide representations of SITS whose temporal dimensions match those of the input SITS. As a result, these methods provide unaligned representations of variable temporal size. Therefore, such representations can not be directly used in traditional machine learning methods for regression or classification tasks. Another study \cite{astruc2024omnisat} proposes to collapse the temporal dimension of the latent representation during training. Nevertheless, the proposed latent representation has also a lower spatial resolution than the input, unsuitable for full resolution mapping.

\textbf{(ii) Learning informative representations from unlabeled data.}  The SSL pre-training strategy used to train a FM should yield meaningful high-level SITS representations. Masked Auto-Encoders (MAE) have been frequently employed for SITS pre-training due to their ease of implementation \cite{yuan21_self_super_pretr_trans_satel,yuan22_sits_former,dumeur-2024-self-super,Tseng2023LightweightPT}. SITS encoders are trained to reconstruct parts of the input SITS. Nevertheless, the ability of MAE to extract high-level features has been questioned \cite{assran-2022-masked-siames}. Indeed, it is hypothesized that being trained to predict in a low-level semantic space (pixel level), could reduce the ability to learn more complex and abstract SITS representations. 
In addition, other SSL training techniques compute a loss at the level of the latent space, which is supposed to be of a higher semantic level than the input data.
For example, instance discrimination strategies are multi-view SSL techniques designed to maximize the similarity between representations of two views from the same input data. Views of the input data must be generated in such a way as to preserve the semantic information needed to solve downstream tasks. While artificial data augmentation are often employed to generate views in computer vision, such techniques are not relevant for SITS. Apart from SkySense \cite{guo2024skysense}, instance discrimination remains largely unexplored in SITS because it requires domain specific view generation, and often needs large batch sizes. 
To mitigate the disadvantages of both types of SSL strategies, recent works propose hybrid approaches \cite{Wang2022RethinkingMS,tsai2021selfsupervised}. These methods combine various SSL strategies, such as instance discrimination with MAE, to learn more informative representations. Except for \cite{astruc2024omnisat}, these hybrid approaches are not applied to SITS representation learning. 

\textbf{(iii) Providing and assessing generic representations.} The genericity of a SITS representation refers to the ability of a model to perform well on two important scenarios: first, in a variety of geographic and temporal configurations, and second, in a variety of downstream tasks.
Under this purpose, it is crucial to  pre-train FM on large-scale data-sets covering diverse geographical and temporal configurations, which avoids learning representations only relevant to specific areas and periods.
Nevertheless, there is currently  a significant lack of well-designed large-scale unlabeled datasets for SITS.  Existing datasets \cite{wang-2023-ssl4eo-s12,satmae2022,Prithvi-100M-preprint} do not provide enough temporal acquisitions or correspond to restricted geographical and temporal configurations \cite{BibEntry2023May}.  In addition, the production of RS FM is also limited by the lack of downstream reference tasks that are representative of the needs of the geosciences and climate communities. While several FM \cite{Tseng2023LightweightPT,satmae2022} in RS are evaluated on scene-classification tasks, most real-world RS applications necessitate high spatial resolution semantic maps. Despite some growth, downstream labeled segmentation datasets for SITS remain scarce, limiting the evaluation of FM in the production of generic spectro-spatio-temporal representations of SITS.

In view of the above (i), (ii), (iii), we propose an aligned SITS encoder (ALISE) as a further step towards the development of a FM for SITS. As the numerous criteria mentioned above already require substantial research, ALISE was designed to process data from just one sensor (S2). Although ALISE is not a FM it addresses  the production of easy-to-use, informative and generic representations. First, ALISE furnishes aligned and fixed-size representations, leveraging the spatial, spectral and temporal dimensions of multi-year SITS, which are easy-to-use (i). The resulting ALISE representations also preserve the spatial resolution of the input SITS, which is considered as crucial for downstream segmentation tasks. Secondly, hybrid SSL strategies are investigated to obtain informative SITS representations (ii). Notably,  we have studied the possibility of integrating an SSL instance discrimination strategy alongside an MAE task. We propose a cross-view reconstruction task, where views are subseries of the original SITS which are each composed of different acquisitions. We also investigate whether integrating additional instance discrimination losses leads to more informative latent representations. These losses enforce invariance between the SITS views representations and decorrelate latent variables. Thirdly, generic representations are sought by using a new pre-training unlabeled open-source dataset and evaluating the model on three downstream tasks (iii). In particular, the three distinct downstream tasks are: crop segmentation (PASTIS \cite{garnot-2021-panop-segmen}), dense land cover segmentation (MultiSenGE \cite{wenger-2022-multis}), change detection (with the specially designed \emph{CropRot} dataset \cite{croprot}). This novel labeled dataset was built to enhance the benchmark of downstream tasks for the assessment of FM. On the two segmentation downstream tasks, we train a single linear layer to perform pixel level classification. The quality of ALISE's representations is evaluated by using them in frozen (linear-probing) and fine-tuning configurations. Finally, the change detection task is performed without any additional learning step. Change maps are generated by measuring the distance between a pair of aligned SITS representations obtained from ALISE. In addition, this paper details: ALISE’s performance under a labeled data scarcity scenario, a qualitative assessment of the proposed temporal alignment method, and an extensive study of the influence of the view generation protocol and instance discrimination losses.

Our contributions can be summarized as follows:
\begin{itemize}
\item We introduce ALISE, a novel SITS encoder that provides aligned representations of SITS at high spatial resolution.
\item We explore a new multi-view SSL task specifically designed for SITS.
\item We provide two novel datasets: a large scale unlabeled pre-training multi-year European S2 dataset \cite{mmdceu} and a labeled downstream crop change detection dataset \cite{croprot}.
\item We achieve state-of-the-art performance on linear probing segmentation tasks \cite{Tseng2023LightweightPT}, \cite{dumeur-2024-self-super}.
\end{itemize}

The code\footnote{\url{https://src.koda.cnrs.fr/iris.dumeur/alise}} and the pre-training \cite{mmdceu} and change detection \cite{croprot} datasets are available.
The remainder of the paper is organized as follows. First,  \autoref{sec:related-works} presents the current state-of-the-art in terms of SSL strategies for SITS, methods for aligning irregular and unaligned SITS and pre-training datasets. Next, \autoref{sec:methods}  corresponds to our proposed methodology. Moreover, the explanation of the experimental setup is detailed in \autoref{sec:experimental-setup} and the results in  \autoref{sec:results}. Finally conclusions are drawn in \autoref{sec:conclusion}.

\section{Related works}
\label{sec:related-works}
In this section we review works related to the production of informative (\autoref{sec:ssl-mae}, \autoref{sec:ssl-id}), ready-to-use (\autoref{sec:rw-align}) and generic (\autoref{sec:rw-datasets}) representations. 
More specifically, this section presents an overview of the two main categories of SSL, which are: MAE applied in most cases with SITS (\autoref{sec:ssl-mae}), and instance discrimination SSL strategies which have rarely been explored in the context of SITS (\autoref{sec:ssl-id}). Subsequently, methodologies for the generation of aligned SITS representations are presented (\autoref{sec:rw-align}), followed by an overview of existing S2 large-scale pre-training datasets (\autoref{sec:rw-datasets}).
\subsection{Masked auto-encoders for time series}
\label{sec:ssl-mae}
Masked auto-encoders (MAE) were popularized thanks to the great performance obtained by BERT \cite{Devlin2018}. The MAE strategy involves corrupting multiple elements (tokens) of the input sequence and training the model to reconstruct them. This SSL strategy was soon extended to other domains such as time series \cite{nie2023a,liu2023pttuning,cheng2023timemae} or image processing \cite{he-2022-masked-autoen}.
In the field of RS, a variety of studies employed MAE in the analysis of mono-date satellite images \cite{rvsa,hong-2024-spect} or satellite video \cite{yao-2023-ringm-sense}.
For SITS, the proposed MAE use either a temporal masking strategy with a temporal transformer or a spatio-temporal masking strategy with a Vision Transformer (ViT).
As detailed in \cite{dumeur-2024-self-super}, existing spatio-temporal masking strategies, such as the SatMAE \cite{satmae2022} and Prithvi \cite{Prithvi-100M-preprint}, process exclusively SITS composed of three acquisitions. The temporal approaches, on the other hand, are often applied on fully-temporal architectures \cite{yuan21_self_super_pretr_trans_satel,Tseng2023LightweightPT} or with narrow spatial context \cite{yuan22_sits_former}. To the best of our knowledge, the sole spatio-spectro-temporal architecture adapted to SITS and pre-trained as an MAE is U-BARN \cite{dumeur-2024-self-super}.
Furthermore, MAE methodologies adapted to time series (not specifically SITS) differ on two important points.

First, they differ on how they handle corrupted tokens. Inspired by the original BERT \cite{Devlin2018}, methods with temporal masking inject the corrupted tokens directly into the SITS encoder. Unfortunately, this input data corruption introduces a distribution shift between pre-training and downstream tasks, as the latter have no input corruption step. Spatio-temporal methods \cite{Prithvi-100M-preprint,satmae2022} address this last limitation by using an asymmetric encoder-decoder architecture. Inspired by MAE in vision \cite{he-2022-masked-autoen}, the corrupted tokens are not fed to the encoder. Instead, the latter are concatenated to the input representations and processed solely by the decoder using a self-attention mechanism. In addition, outside RS studies, recent MAE for time series such as \cite{liu2023pttuning,cheng2023timemae} also employ an asymmetric architecture. Instead of a self-attention mechanism, a lightweight decoder that performs cross-attention between corrupted tokens and the latent representation, is employed.

Secondly, MAE on time series differ in the employed masking pattern. While retrieving a masked word in natural language processing (NLP) requires a holistic understanding of the sentence, neighboring data points in time series are often highly correlated. Therefore, several studies \cite{cheng2023timemae,liu2023pttuning,nie2023a} advocate splitting the time series into non-overlapping temporal sub-series before masking. Therefore, the masking strategy is applied at the sub-series level to force the model to reconstruct local variations. However, this methodology is not directly applicable to irregular SITS, where each sub-series would represent different temporal scales. Existing methods on SITS, thus often mask random time steps \cite{dumeur-2024-self-super,yuan22_sits_former}, ignoring the potential redundancy of the temporal information. Sole \cite{Tseng2023LightweightPT} considers consecutive acquisition masking.

Consequently, unlike several previous studies on SITS \cite{yuan21_self_super_pretr_trans_satel,yuan22_sits_former,dumeur-2024-self-super,satmae2022}, we propose a temporal masking strategy, where the corrupted tokens are not processed by the SITS encoder. We also consider masking successive acquisitions, and the reconstruction task utilizes a lightweight decoder with cross-attention.

\subsection{Self-supervised instance discrimination methods}
\label{sec:ssl-id}
The use of MAE in vision has nevertheless been criticized for focusing on learning local relationships within an input sample, instead of modeling the relationship between samples \cite{huang2023contrastive}. Besides, while MAE techniques are easy to implement, they can produce representations that are generally of a lower semantic level than instance discrimination techniques \cite{assran2023self}.

As per \cite{tao2023siamese}, we consider instance discrimination as a subset of SSL, where a siamese network (composed of two branches) is trained to produce similar representations of two views of the same data. The views correspond to alternative ways of observing the input data.  Different views can be, for example, signals from different sensors or artificially augmented versions of the input data. In addition, views are expected to preserve the input semantic information needed for downstream tasks.
While computer vision studies often employ a set of benchmark data augmentation techniques, determining relevant view generation on SITS is challenging.
Nevertheless, several studies have tackled view generation methods on single RS image analysis. For instance, \cite{manas-2021-season-contr} have shown the interest of using domain-adapted augmentation on RS images. These methods propose to consider two images taken at the same location but at different times as two views of the same input data. Other works exploit the multi-modality and consider images from different sensors (optical and radar pairs) as two different views \cite{jain2022self,wang2022self}. 

Multi-view SSL techniques can be divided into four categories: contrastive \cite{pmlr-v119-chen20j}, clustering  \cite{caron-2018-deep-clust,10.5555/3495724.3496555},  distillation \cite{caron:hal-03323359,NEURIPS2020_f3ada80d,9578004} and redundancy reduction \cite{pmlr-v139-zbontar21a,bardes2022vicreg,bardes2022vicregl}. These approaches differ in their strategies to prevent representation collapse, a scenario in which the encoder always predicts the same representation regardless of the input.

Contrastive learning \cite{pmlr-v119-chen20j} and its variants for segmentation tasks \cite{10.5555/3495724.3496101} heavily rely on negative pair sampling (i.e. finding pairs of samples representing different semantics). Efficient negative pair sampling is challenging for SITS because pixels from different SITS may still represent the same classes. In the absence of labels, it is difficult to identify negative examples that are not trivial and are actually beneficial to learning. Consequently, for pixel-level SITS classification, a contrastive loss is often used in a semi-supervised framework where labels help generating relevant negative pairs \cite{yuan-2023-bridg-optic}.
SSL clustering methods, on the other hand, do not require sampling of negative pairs. Instead, they use a clustering algorithm dedicated to the generation of pseudo-labels (also called prototypes). To do so, strong assumptions about the batch distribution are required. For example, \cite{10.5555/3495724.3496555} assumes that all examples in a batch are evenly distributed among the prototypes. Distillation-based techniques \cite{caron:hal-03323359,NEURIPS2020_f3ada80d,9578004}, on the other hand, require complex training tricks such as momentum-encoder or stop-gradient. 
Lastly, compared to other instance discrimination SSL frameworks, the implementation of redundancy reduction techniques \cite{pmlr-v139-zbontar21a,bardes2022vicreg,bardes2022vicregl} is straightforward. These strategies prevent informational collapse by decorrelating every pair of variables of the embedded latent representation. The VicReg approach \cite{bardes2022vicreg} proposes the use of three losses: the invariance loss, which enforces similarity between the embedded latent representations of the two views; the variance loss, which maintains the variance of the embedded variables above a threshold; and the co-variance loss, which intends to decorrelate the variables of each embedded view. Furthermore, a modified version of VicReg, named VicRegL \cite{bardes2022vicregl}, has been adjusted for downstream segmentation tasks. In this latter method the three previous losses are also calculated at the pixel level.
Consequently, due to its simplicity, the combination of VicReg losses with a cross-reconstruction multi-view task is explored in this paper.
\subsection{Processing irregular and unaligned SITS}
\label{sec:rw-align}
For downstream tasks, SITS representations provided by pre-trained models must be of fixed-size, in order to be injected into classical lightweight classifiers including Random Forest \cite{inglada-2017-operat-high}, TempCNN \cite{pelletier-2019-tempor-convol}, Recurrent Neural Networks \cite{ienco2017land}, Sparse Variational Gaussian Process (SVGP) \cite{bellet2023land}. Additionally, those representations should be "aligned", enabling the direct comparison of the features of two different samples.

The generation of aligned SITS prior to their use in a machine learning (ML) model has been addressed in several works. A common practice is to perform a linear interpolation of the annual time series into a common temporal grid \cite{inglada-2017-operat-high}. While this method is efficient for annual SITS classification, it might remove fine-grain information or introduce noise. Besides, cloud masks are required to perform the interpolation, and the definition of reference dates, composing the common temporal grid, is challenging.
Another common pre-processing method is the generation of \emph{composite} images, which aims to summarize valid acquisitions over a temporal extent. For example, the authors of Presto \cite{Tseng2023LightweightPT} suggest using time series with monthly information to fuse multi-sensor SITS. In this case, optical SITS monthly information corresponds to the least cloudy image, while a median of SAR acquisitions over a month is used. Unfortunately, this temporal pre-processing strategy to prevent irregular temporal feature representations has several shortcomings. First, the proposed monthly optical sampling protocol does not ensure that each pixel of the image has a clear acquisition. Second, the monthly sampling protocol induces an important loss of temporal information.

Furthermore, the latter two methods are not flexible and do not adjust to the target task. For this reason "data-driven alignment strategies" have been proposed. For instance, \cite{10363355} proposes the learning of an attention-based interpolation (mTAN) \cite{shukla2021multitime}, which is trained end-to-end along a Gaussian Process classifier (SVGP). In particular, the interpolation weights are computed thanks to a scaled dot product between embedded acquisition dates and embedded reference dates. The temporal embedding of the dates is performed with learnable embedding functions \cite{kazemi2020timevec}. Besides, \cite{10363355} shows that the attention-based interpolation framework outperforms SVGP fed with linearly interpolated data. Although this mechanism is more flexible and relevant for a classifier requiring aligned SITS such as SVGP, novel attention based encoder architectures can now process irregular and unaligned SITS \cite{russwurm20_self_atten_raw_optic_satel}. With these latter networks, a temporal pre-processing will result in an unnecessary loss of information. As a consequence, aligning SITS after their encoding by an attention-based encoder, seems a more appropriate approach. A second limitation of mTAN is that the interpolation weights are calculated exclusively with the acquisition dates, without consideration of the content of the time series. 

Recently, in computer vision, advanced DL architectures such as the Perceiver I/O \cite{jaegle2022perceiver} have emerged to map arbitrary input sequences onto a fixed-size aligned latent space using a flexible cross-attention querying mechanism. Specifically, the projection of the input sequence is determined by a scaled dot product between the input sequence and a learnable array, denoted \emph{learnable queries}. The resulting length of the output aligned sequence is determined by the number of learned queries. This mechanism has been applied to SITS with a single learnable query in \cite{astruc2024omnisat} to provide SITS representations with a collapsed temporal dimension. However, the use of this mechanism with a larger number of queries and its analysis has not yet been investigated.

As a result, we propose a flexible query mechanism to align irregular and unaligned SITS after their encoding by an attention based encoder \cite{Vaswani2017}. We also investigate how the information is stored in the aligned latent representations.

\subsection{Sentinel-2 pre-training data-sets}
\label{sec:rw-datasets}
As the number of attempts to construct a RS FM increases, a multitude of pre-training data-sets have been constructed. \autoref{tab:pretraining-datasets} provides an overview of existing large scale pre-training datasets employed to pre-train SITS encoders. Ideally, a RS FM should be trained on multi-spectral data covering multiple geographic and temporal configurations. The pre-training data should also include SITS with numerous acquisitions to allow learning complex temporal features. Moreover, a SITS FM is expected to learn to ignore cloudy pixels. As a consequence, invalid pixels should not be removed from the pre-training input data. Finally, to train a spatio-spectro-temporal network, it is necessary to collect SITS with a large spatial extent.

With the exception of the Prithvi data-set, all the proposed data-sets provide at least the ten S2 bands (10m-20m resolution) bands. In addition, except for FR-S2, the data-sets show remarkable geographical variability. However, none of the existing data-sets fits all the aforementioned criteria to pre-train a large-scale SITS FM. Indeed, most of these data-sets \cite{wang-2023-ssl4eo-s12,Prithvi-100M-preprint,Tseng2023LightweightPT} do not provide long enough time series to capture the complex temporal dynamics of the Earth's surface. 
Additionally, several pre-training data-sets employ a severe cloud filtering of the information (SSL4EO-S12, Presto and Clay data-sets). As a consequence, this filtering prevents the FM from identifying cloudy pixels as irrelevant. This could decrease performance on downstream tasks where the cloud mask information may not be available. In contrast, a pre-training data-set like FR-S2 applies less strict cloud filtering, and provides validity mask.
Besides, it is presumed that a RS FM pre-training dataset should be balanced \cite{Zhu2024May}. Nevertheless, some existing datasets, such as SSL4EO-S12 \cite{wang-2023-ssl4eo-s12}, focus on urban areas. The scarcity of natural landscapes in such datasets could be an obstacle to learning complex temporal dynamics.

Consequently, we introduce a novel large-scale European S2 SITS pre-training data-set named MMDC-EU, which is detailed in \autoref{sec:pre-training-datasets}. 
\begin{table*}[htbp]
\caption{\label{tab:pretraining-datasets}Description of S2 SITS unlabeled training data-sets used to pre-train large scale SITS models. Temporal extent refers to the longest time interval existing in the data-set. The number of dates corresponds to the exact or average (\(\sim\)) number of dates in the SITS. The ROI size corresponds the spatial dimensions of the images within the SITS. “Available” indicates whether the data-set is available for download. “Cloud filter” indicates whether cloudy images have been removed. When the percentage is shown, it specifies the maximum percentage of clouds accepted in an image. "\textbf{?}" is employed when the information is not given in the corresponding article.}
\centering
\begin{tabular}{|p{0.15\linewidth} |p{0.10\linewidth}|p{0.05\linewidth}|p{0.07\linewidth}|p{0.1\linewidth} |p{0.1\linewidth}|p{0.08\linewidth}|p{0.05\linewidth}|p{0.1\linewidth}|}
\hline
Data-Set Name & Data & Temporal Extent & Number of dates & Geographical extent & Roi size & Available & Cloud filter\\
\hline
\hline
SSL4EO-S12 \cite{wang-2023-ssl4eo-s12} & S2-L2A 13 bands & 2020 & 4, (1/season) & Worldwide & \(264 \times 264\) & \ding{51} & yes, \(\leq 10\) \%\\
\hline
FR-S2 (U-BARN)  \cite{dumeur-2024-self-super} & S2-L2A 10 bands + valid mask & 2018-2020 & \(\sim 179\) & France & \(1024 \times 1024\) & \ding{51} & yes, \(\leq 30\) \%\\
\hline
Presto \cite{Tseng2023LightweightPT} & S2-L2A 10 bands & 2020-2021 & 24 (1/month) & Worldwide & \(1 \times 1\) & \xmark & yes\\
\hline
Prithvi \cite{Prithvi-100M-preprint} & NASA HLS\footnotemark V2 L30 & \textbf{?} & \textbf{?} & USA & \(64 \times 64\) & \xmark & yes\\
\hline
SkySense \cite{guo2024skysense} & S2 L2A 10 bands & \textbf{?} & \(\sim 65\) & Wordlwide & \(64 \times \times 64\) & \xmark & yes, \(\leq 1\) \%\\
\hline
Clay\footnotemark & S2 L2A  10 bands & 2018-2023 & 8, (1/quarter) & Worldwide & \(224 \times 224\) & \ding{51} & yes\\
\hline
MMDC EU (ours) & S2 L2A 10 bands + valid mask & 2017-2020 & \(\sim 354\) & Europe & \(512 \times 512\) & \ding{51} & no\\
\hline
\end{tabular}
\end{table*}\footnotetext[2]{\label{org53c0469}Harmonized LandSat Sentinel 2 data-set}\footnotetext[3]{\label{orgbd8304e}\url{https://clay-foundation.github.io/model/release-notes/data\_sampling.html}}

\section{Method}
\label{sec:methods}
The method consists in pre-training an ALIgned SITS Encoder, ALISE, which produces aligned representations for multi-year irregular and unaligned SITS. The details of the ALISE architecture are presented in the next section, followed by the description of the multi-view self-supervised pre-training strategy. Specifically, the proposed SSL method studies the combination of two types of losses: a generative cross-reconstruction loss and instance discrimination losses computed in the latent space. 

\subsection{ALISE: ALigned SITS Encoder}
\label{sec:alise}
ALISE harnesses the spectral, spatial, and temporal dimensions of irregular and unaligned input time series \(X \in \mathbb{R}^{(b_s,t,c,h,w)}\), where \(b_s,t,c\) are respectively the batch, temporal, and spectral dimensions and \(h,w\) the spatial dimensions. Although \(t\) may vary for each SITS, ALISE generates a latent representation \(Y \in \mathbb{R}^{(b_s,n_q,d_{model},h,w)}\) of fixed dimensions, where \(d_{model}\) and \(n_q\) are the channel and temporal sizes.

\begin{figure}[htb]
\centering
\includegraphics[width=.9\linewidth]{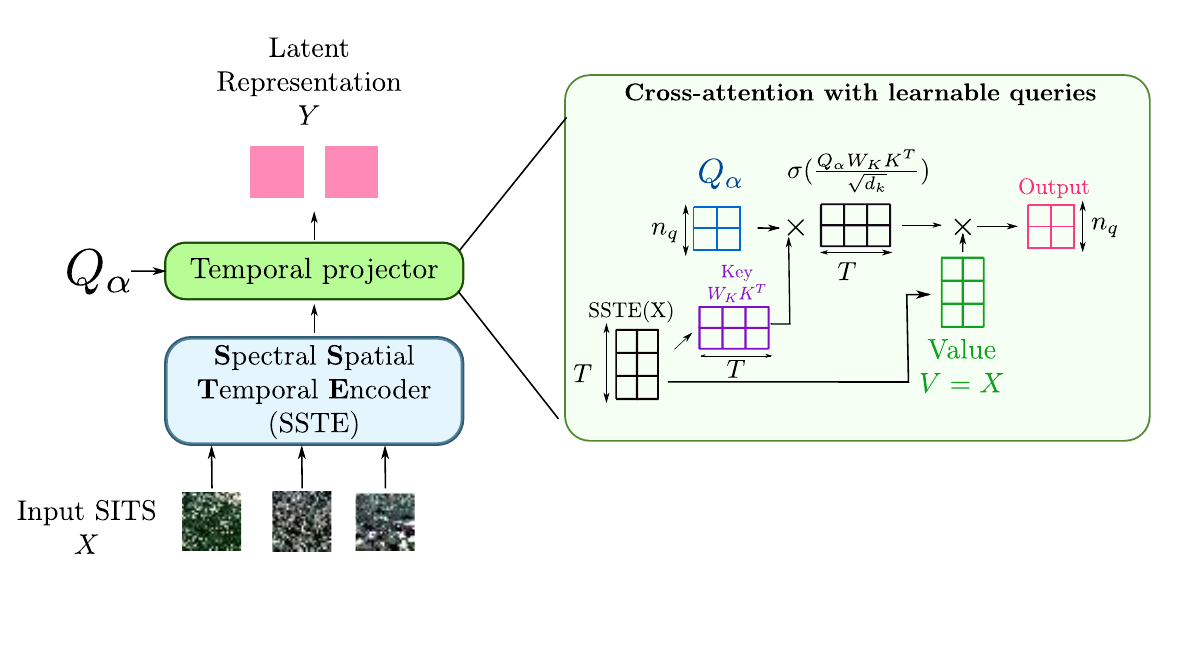}
\caption{\label{fig:alise-archi}Overall description of ALISE architecture. The input time series X is first processed by the spectral spatial temporal encoder (SSTE) \cite{dumeur-2024-self-super}. The obtained intermediate representations are then processed by a temporal projector. The temporal projector corresponds to a cross-attention mechanism with learnable queries \(Q_{\alpha}\). For visual clarity, the cross-attention is represented for one attention head.}
\end{figure}
As illustrated in \autoref{fig:alise-archi}, ALISE is composed of two main blocks. First, a Spatial, Spectral and Temporal Encoder (SSTE), noted \(\Psi\) in \Cref{eq:sdp-lq}, which corresponds to the U-BARN architecture detailed in \cite{dumeur-2024-self-super}. As the original U-BARN was initially designed to handle annual SITS, the positional encoding (PE) in ALISE has been modified to process multi-year SITS.  Specifically, the temporal information provided to ALISE is not the Day of Year (DoY). Instead, \(\delta_t\), which is the difference in days between the image acquisition date and a given reference date (03/03/2014), is employed.
\begin{equation}\label{eq:sdp-lq}
O^h=\sigma \left( \frac{Q_{\alpha^h} W_1^{h^T} \Psi(X)^T}{\sqrt{d_{model/H}}} \right) \Psi(X)
\end{equation}
Besides, a temporal projector processes the irregular and unaligned outputs of the SSTE, \(\Psi(X)\), to generate aligned SITS representations. The projection is performed by a multi-head temporal cross-attention mechanism between learnable queries and \(\Psi(X)\).  The scaled dot product of the cross-attention mechanism on a head \(h\) is detailed in \Cref{eq:sdp-lq} with \(Q^h_{\alpha}\) the learnable queries, \(X\) the input time series, \(d_{model}\) the number of features in \(\Psi(X)\), H the number of heads, and \(\sigma\) the softmax function. The attention product is fully temporal, thus \(Q^h_{\alpha} \in \mathbb{R}^{(n_q,d_{model/H})}\), \(W_1^h \in \mathbb{R}^{(d_{model},d_{model}/H)}\) and  \(\Psi(X) \in \mathbb{R}^{(t,d_{model})}\). As detailed in \autoref{eq:lat-repr} , all the outputs of each head \(O^h \in \mathbb{R}^{n_q,d_{model}/H}\) are then concatenated along the feature dimension and processed by a Multi-Layer Perceptron (MLP), which generates the latent representation \(Y\).
\begin{equation}\label{eq:lat-repr}
Y=\text{MLP}(\text{concat}_h \left(O^1, \ldots, O^H \right))
\end{equation}
The temporal dimension of the latent representation \(Y\) is determined by the number of learnable queries (\(n_q\)). It must be noted that the temporal projector does not shrink the spatial dimension of \(\Psi(X)\), meaning that each pixel of the SITS is represented by \(d_{model}\) features along \(n_q\) temporal features.
It is crucial to understand that in the resulting latent representations the notion of "time" is not preserved. In other words, the time series is folded in a way that does not preserve the notion of order or distance in the time axis.  As a result, the latent representation is not a time series; rather, it is a stack of temporal features. Therefore, we refer to the \(n_q\) vectors in the aligned latent representation as the "latent temporal features". 

\subsection{Multi-view pre-training task}
\label{sec:orgce2b8e6}
\begin{figure}[htb]
\centering
\includegraphics[width=.9\linewidth]{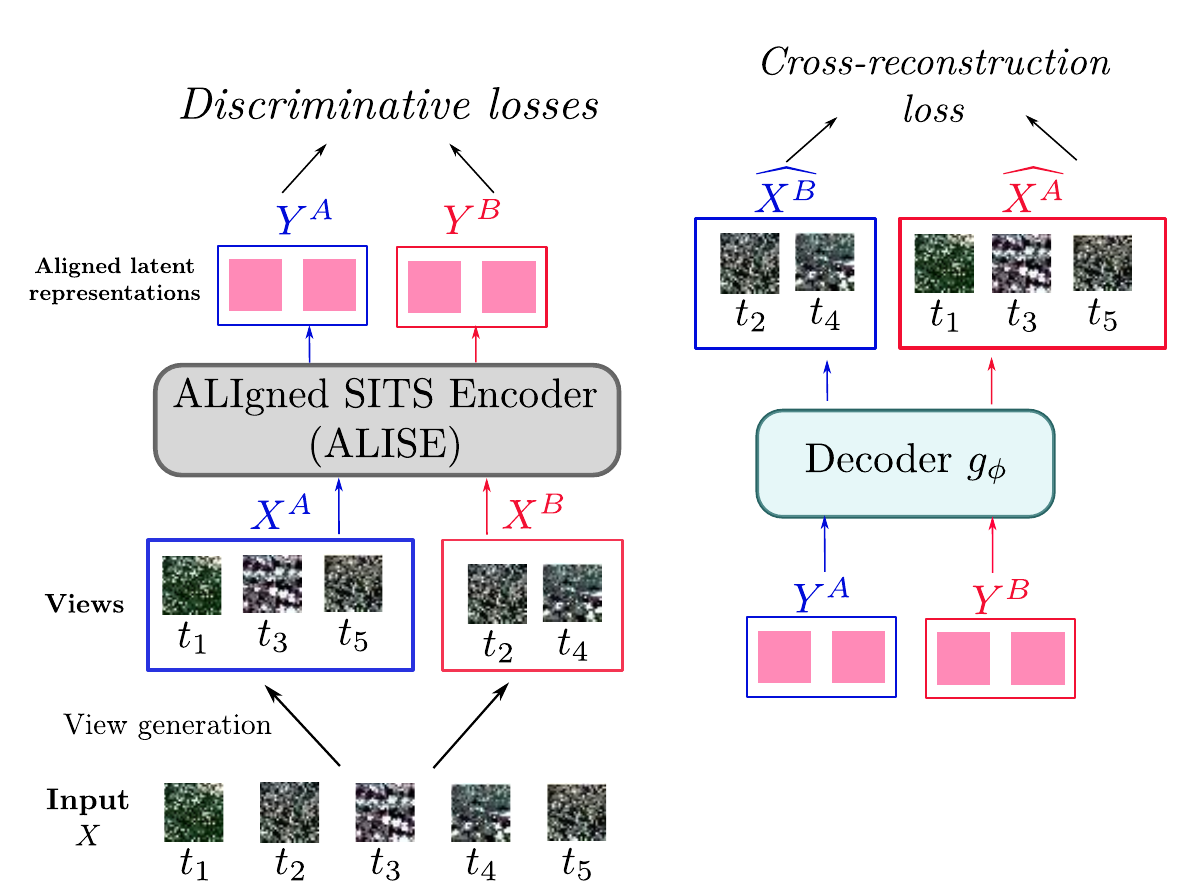}
\caption{\label{fig:mvssl-overall}Description of the proposed multi-view SSL strategy. Given an input time series \(X\) two views are generated: \(X^A\) and \(X^B\). Each view is processed independently by ALISE which generates their respective aligned latent representations \(Y^A\) and \(Y^B\). A decoder \(g_{\phi}\) is trained to reconstruct one view using the latent representation of the other. Additional discriminative losses are computed on the latent representation.}
\end{figure}

As detailed in \autoref{fig:mvssl-overall}, the multi-view SSL pre-training task, combines a cross-reconstruction loss with two additional losses computed on the embedded latent representations.
As detailed in \Cref{glob:loss}, the total SSL loss, corresponds to the weighted sum of three terms \(L_{inv}\), \(L_{cov}\) and \(L_{rec}\) respectively the invariance, covariance and reconstruction losses.
\begin{equation} \label{glob:loss}
L= w_{inv} L_{inv} + w_{cov} L_{cov} + w_{rec} L_{rec}
\end{equation}
\subsubsection{View generation}
\label{sec:view-gene}
After the view generation phase, ALISE encodes each of the two views, resulting in two aligned representations. The latter are used to compute the invariance and covariance losses. In the cross-reconstruction loss, the representation of each view is used to reconstruct the other view.
The objective of the view generation protocol is to provide views that preserve semantic content. For SITS, the generation process aims to create views that maintain the pixel information of the observed landscape. Consequently, the two views, \(X^A\) and \(X^B\), represent a time series at the same location but with different acquisition times. The view generation process starts by selecting \(N\) adjacent acquisitions among an irregular and multi-year SITS. As detailed in \Cref{eq:idx-view-global}, this latter time series is divided into \(n_w\) non-overlapping temporal windows, each composed of \(t_w\) dates. Given that SITS are irregular, each temporal window may represent a different temporal span.
\begin{equation} \label{eq:idx-view-global}
X = \bigcup_{i=0}^{n_w-1}\left \{ X_j \mid {i \times t_w}\leq j < {(i+1) \times t_w} \right \}
\end{equation}
Finally, to ensure that the two views cover nearly identical periods, every other temporal window is used to construct respectively \(X^A\) (\Cref{eq:idx-viewa}) and \(X^B\) (\Cref{eq:idx-viewb}). Therefore, \(t_w\) corresponds to the number of consecutive dates that the model has to predict during the training process. We posit that increasing \(t_w\) complexifies the cross-reconstruction task as more temporal variations have to be retrieved by the model. This generation approach ensures that the views are temporally intertwined: \(X^A \cup X^B = X\) and \(X^A \cap X^B = \emptyset\) and provides a parameter \(t_w\) which controls the difficulty of the pre-training task. 

\begin{subequations}
\begin{align}
X^A = \bigcup_{i=0}^{\frac{n_w}{2}-1 }\left \{ X_j \mid {2 \times i \times t_w}\leq j < {(2 \times i+1) \times t_w} \right \} \label{eq:idx-viewa}\\
X^B = \bigcup_{i=0}^{\frac{n_w}{2}-1 }\left \{ X_j \mid {(2 \times i+1) \times t_w} \leq j < {(2 \times i+2) \times t_w} \right \} \label{eq:idx-viewb}
\end{align}
\end{subequations}
\subsubsection{Discriminative losses}
\label{sec:latentlosses}
\begin{figure}[htb]
\centering
\includegraphics[width=0.4\linewidth]{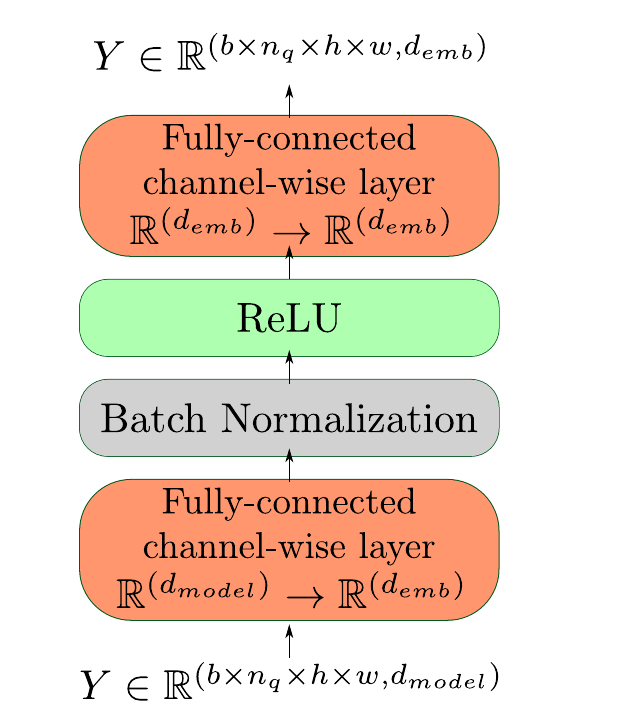}
\caption{\label{fig:projector-archi}Description of the projector architecture.}
\end{figure}

As illustrated in \Cref{fig:mvssl-overall}, the augmented views \(X^A\), \(X^B\) are independently encoded by ALISE. The aligned latent representations \(Y^A\) and \(Y^B\) are then processed into embeddings \(Z^A\), \(Z^B\) by a projector. The projector aims to eliminate the information by which the two representations differ. Specifically, the projector operates exclusively on the channel dimensions: \(\pi_{\omega}:\mathbb{R}^{( d_{model})} \rightarrow \mathbb{R}^{( d_{emb})}\). In other words, pixel-level latent vectors of each \(n_q\) query are independently processed by the projector. As shown in \Cref{fig:projector-archi}, the proposed projector consists of one fully connected layer followed by batch normalization and ReLU, and a second linear layer. It is assumed that the choice of the projector's architecture affects the computation of the covariance loss. However, no empirical benefits were found from using a deeper or wider projector architecture for our considered downstream tasks.

We denote \(\mathbf{z}^k_{(b,n,i,j)} \in \mathbb{R}^{d_{emb}}\) the pixel-level embedded vector of \(Z^k\) located at the spatial position (i,j) for the  \(n\)\textsuperscript{th} query and \(b\)\textsuperscript{th}  batch position. We propose to compute the invariance and covariance losses on the embeddings \(Z^A\) and \(Z^B\).

First, the invariance loss maximizes the similarity between the embedded vectors \(\mathbf{z}^A\) and \(\mathbf{z}^B\) (see \Cref{eq:invar-loss}).  As \(X^A\) and \(X^B\) have distinct acquisition dates but cover the same time-period, \(L_{inv}\) aims at learning representations which are invariant to the acquisition dates.

\begin{equation}\label{eq:invar-loss}
L_{inv}(Z^A,Z^B)=\frac{1}{b_s \times n_q \times h \times w}\sum_{(b,n,i,j)} \lVert \mathbf{z}^A_{b,n,i,j}- \mathbf{z}^B_{b,n,i,j} \rVert_2^2
\end{equation}

Second, we also investigate whether the covariance loss allows learning better representations. The covariance loss decorrelates the different \(d_{emb}\) features. The total covariance loss, \Cref{eq:cov}, corresponds to the sum of the covariance losses computed for each embedding \(Z^k\). For centered embeddings \(Z \in \mathbb{R}^{(b_s \times n_q \times h \times w,d_{emb})}\) the covariance loss aims to minimize the off-diagonal values of the co-variance matrix \(C(Z)\) in \Cref{eq:covmat}. In other words, the covariance matrix of the \(d_{emb}\) variables, is estimated on a batch composed of \(b_s \times n_q \times h \times w\) samples. In \Cref{app:vicregl}, we discuss how these discriminative losses are related to the VicRegL \cite{bardes2022vicregl} losses. 

\begin{equation}\label{eq:covmat}
l_{cov}(Z)=\frac{1}{d_{emb}}\sum_{i \neq j}[C(Z)]^2_{i,j}
\end{equation}
\begin{equation}\label{eq:cov}
L_{cov}=l_{cov}(Z^A)+l_{cov}(Z^B)
\end{equation}
\subsubsection{Cross reconstruction task}
\label{sec:crossrec}
\begin{figure}[htb]
\centering
\includegraphics[width=.9\linewidth]{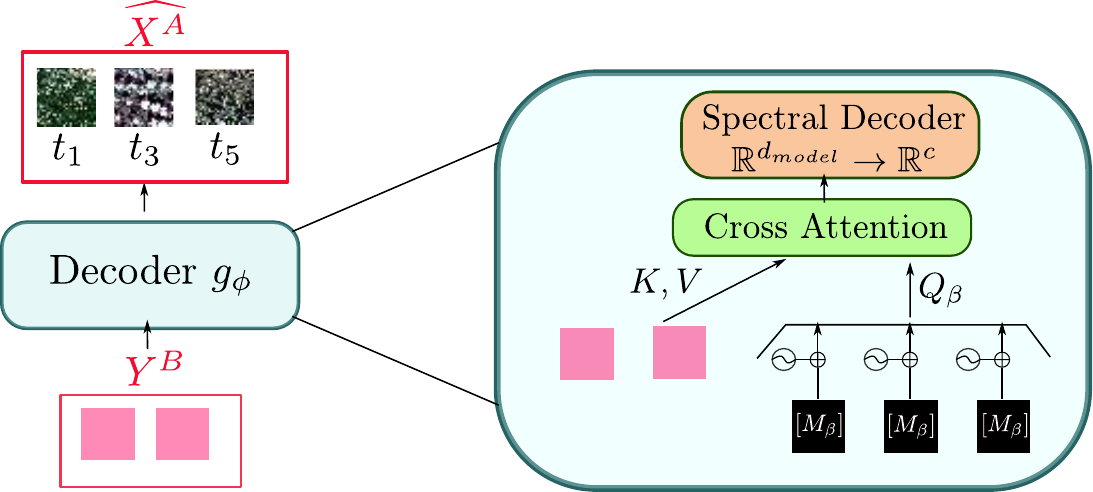}
\caption{\label{fig:cross-attndec}Description of the lightweight decoder employed for the cross-reconstruction task.}
\end{figure}

As depicted in \autoref{fig:cross-attndec}, the latent representations \(Y^A\), \(Y^B\) are employed in a cross-reconstruction task.
A specific fully-temporal decoder using a cross-attention mechanism followed by a fully-connected layer is trained to recover one view \(X^B\) (resp. \(X^A\)) from the latent representation \(Y^A\) (resp. \(Y^B\)) of the other view \(X^A\) (resp. \(X^B\)). The fully-connected layer operates exclusively on the channel dimension of each pixel of the images, to recover the S2 bands from the \(d_{model}\) features.

As proposed in \cite{liu2023pttuning} the cross-attention mechanism exploits \(Q_{\beta} \in \mathbb{R}^{(t_w \times n_w,d_{model})}\) which specifies the dates to be reconstructed.
As detailed in \Cref{eq:masked-queries}, \(Q_{\beta}\) corresponds to the sum of a shared learnable masked token \(M_{\beta} \in \mathbb{R}^{(d_{model})}\) with the temporal positional encoding\footnote{The temporal positional encoding used is the same as the one employed in ALISE.} of the date to reconstruct. Additionally, as described in \Cref{eq:ca}, the latent representation \(Y^k\) with \(k \in \{A,B\}\), is used to construct the keys \(Y^kW_2\) and the values \(Y^k\).
\begin{equation}\label{eq:masked-queries}
Q_{\beta}=\left[ M_{\beta}+PE(\delta_{t_i})\right]_{1 \leq i\leq t_w \times n_w}
\end{equation}
\begin{equation}\label{eq:ca}
\textrm{Cross Attention}(Q_{\beta},Y^k)=\sigma \left(\frac{Q_{\beta} W_1 W_2^T Y^{k^T}}{\sqrt{d_{model}}}\right)Y^k
\end{equation}
Finally, the quality of the reconstruction is assessed by using the classical Mean Square Error.  As described in \Cref{eq:rec}, the reconstruction loss is the average of the reconstruction losses of each view.
\begin{equation} \label{eq:rec}
L_{rec}=\frac{1}{2}[l_{rec}(X^A,Y^B)+l_{rec}(X^B,Y^A)]
\end{equation}
Following the approach of \cite{dumeur-2024-self-super}, pixels with invalid measurements due to the acquisition conditions (e.g. cloudy and out of swath pixels) are ignored in the reconstruction loss. As detailed in  \Cref{eq:one-mse} \(M^{valid}_{t}\) represents the boolean validity mask, \(n^{valid}_{t}\) represents the number of clear pixels, \(T=\frac{n_w \times t_w}{2}\) the number of acquisitions in a view, and \(\odot\) is the Hadamard product.
\begin{equation}\label{eq:one-mse}
l_{rec}(X^k,Y^l)=\frac{1}{T}\sum_{t=1}^T\frac{M^{valid}_{t}}{n^{valid}_{t}}\odot||X_t^k - g_{\phi}(Y^l)_t ||_2^2
\end{equation}
The validity mask is only used in the cross-reconstruction loss and is not included in the input data injected to ALISE. Therefore, no validity masks are required for downstream tasks.
\subsection{Implementation details}
\label{sec:orga354cc1}
To pre-train ALISE, the cosine annealing scheduler with warm restarts \cite{loshchilov2017sgdr} was employed with T\textsubscript{0}=2, and maximum learning rate of 1e-3. To generate the different views from a multiyear SITS, 60 consecutive dates were randomly selected among the 4 years of data. Within our unlabeled data-set, 60 consecutive acquisitions can extend over a maximum of four years of data and a minimum of four months. To increase the diversity of the training data, the selection of the \(t\) consecutive dates used in the view generation is random for each SITS and changes at each epoch. Besides, ALISE architectural hyper-parameters are also detailed in  \Cref{app:alise-archi}. The pre-trainings tasks were conducted on a single Tesla V100 GPU for 260 epochs. The values of the pre-training hyper-parameters are shown in \autoref{tab:def-ssl-hp} and their choices are explained in \autoref{sec:ablation-study}. The pre-trained model with the lowest loss on the pre-training validation set is selected for downstream task assessment.
\begin{table}[h!]
\caption{\label{tab:def-ssl-hp}Default hyper-parameters for pre-training ALISE.}
\centering
\begin{tabular}{|l|l|l|l|l|l|l|l|l|}
\hline
\(t_w\) & \(n_q\) & batch size & \(d_{model}\) & \(d_{emb}\) & w\textsubscript{rec} & w\textsubscript{inv} & w\textsubscript{cov} & H\\
\hline
2 & 10 & 2 & 64 & 128 & 1 & 1 & 0 & 2\\
\hline
\end{tabular}
\end{table}

\section{Experimental setup}
\label{sec:experimental-setup}
First, the four S2 L2A data-sets used in our different experiments are presented: the novel unlabeled large scale data-set (MMDC-EU) used for pre-training ALISE and the three downstream
labeled data-sets (PASTIS, MultiSenGE and the novel \emph{CropRot}). Secondly, the implementation details of our two types of downstream tasks setup (semantic segmentation and change detection) as well as the corresponding competitive works are described.

\subsection{Data-Sets}
\label{sec:org8b6668c}
\begin{figure}[htb]
\centering
\includegraphics[width=\linewidth]{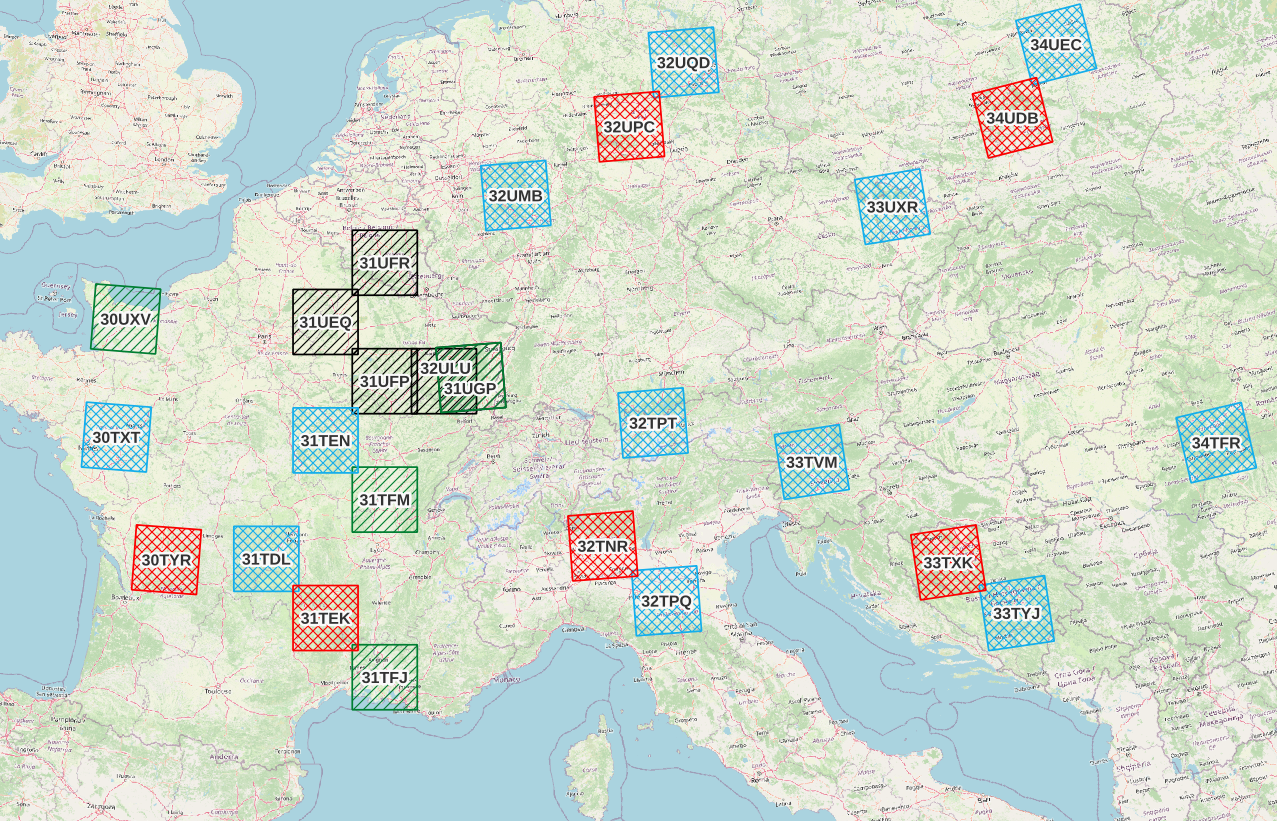}
\caption{\label{fig:datasets}Geographical distributions of the different tiles composing the data-sets. The unlabeled pre-training data-set is composed of multi-year SITS selected within the blue and red boxes for the training and validation data-set respectively. MultiSenGE labeled data are selected in the area delineated by the black boxes. The PASTIS as well as CropRot data-sets are within the green boxes.}
\end{figure}

ALISE is pre-trained on a large scale multi-year European data-set. Besides, three labeled data-sets are used to assess the quality of the pre-trained SITS encoders. The geographical distribution of the different used data-sets is presented in \Cref{fig:datasets}.  For these data-sets, only the four 10 m and the six 20 m resolution bands of S2 are used. The 20 m resolution bands are re-sampled onto the 10 m resolution grid by bi-cubic interpolation. Similarly to \cite{dumeur-2024-self-super}, a robust data normalization is applied on S2 L2A reflectances. 
Due to GPU memory limitations, ALISE is trained to process SITS with a spatial dimension of \(64 \times 64\) pixels (at 10 m resolution). If the used data-set provides larger images, a random crop \footnote{\url{https://pytorch.org/vision/main/generated/torchvision.transforms.RandomCrop.html}} (resp. center crop) is operated during training (resp. validation/testing) steps. 
\subsubsection{MMDC-EU}
\label{sec:pre-training-datasets}
We have constructed an unlabeled, multi-year, multimodal SITS data-set spanning Europe. This multimodal datacube is designated as MMDC-EU. In practice, this data-set is composed of the following data: the S2 L2A product, the Sentinel-1 (S1) ascending and descending acquisitions, ECMWF AGERA5\footnote{\url{https://cds.climate.copernicus.eu/cdsapp\#!/dataset/sis-agrometeorological-indicators?tab=overview}} weather variables, and the Copernicus 30 digital elevation model (DEM). Each SITS of each modality is spatially re-sampled onto the S2 grid. The data cube was downloaded with the openEO platform\footnote{\url{https://openeo.cloud/}}. The code\footnote{\url{https://gitlab.cesbio.omp.eu/dumeuri/openeo\_datasets.git}} used to create the multimodal data-set is provided for reference, allowing for potential future expansion. As this paper proposes a model that processes exclusively S2 SITS, we focus on the description of the pre-training data for this sole modality.
Multi-year S2 SITS from January 2017 to December 2020 were built using all the available acquisitions. The data is split into training and validation sets with respectively 1920 and 180 SITS of spatial dimension \(64 \times 64\) pixels. The downloaded S2 SITS correspond to data processed by Sen2Cor \cite{louis2016sentinel}. The validity mask employed in the cross-reconstruction task is built thanks to the information provided by SLC and CLM layers\footnote{\url{https://docs.sentinel-hub.com/api/latest/data/sentinel-2-l2a/}}.
Specifically, as shown in \autoref{fig:datasets}, the pre-training data-set gathers data from 18 S2 tiles. To build the training data-set, 10 smaller regions of interest (ROIs) of size \(512 \times 512\) pixels are randomly selected from each of the 12 training tiles. The disjoint validation data-set is composed of the remaining 6 S2 tiles, from which 30 ROIs of size \(128 \times 128\) pixels are randomly drawn.

\subsubsection{PASTIS crop segmentation}
\label{sec:orga45dcba}
The PASTIS data-set \cite{garnot-2021-panop-segmen} provides labels for 18 crop classes from the French Land Parcel information System.  The SITS considered in our experiments are collected from January to December 2019. The complete data-set contains 2433 SITS and it is divided into 5 stratified folds. In line with \cite{dumeur-2024-self-super}, the segmentation task is performed exclusively on known crop classes. Background and void classes are ignored. The competitive method Presto requires cloud masks. As these data are not available in the original PASTIS data-set, the raw S2 L2A and their cloud masks were downloaded from the Sentinel hub collection \footnote{\url{https://www.sentinel-hub.com/}}. These S2 data also pre-processed by Sen2Cor are used to assess not exclusively Presto but all models. 

\subsubsection{MultiSenGE land cover segmentation}
\label{sec:org6829209}
MultiSenGE \cite{wenger-2022-multis} is a dense land cover labeled data-set for eastern France in 2020. It is composed of 5 urban classes and 9 natural classes. This data-set is composed solely of images with less than 10\% cloud cover and no cloud mask is provided. The resulting SITS are composed of 3 to 14 acquisitions. In contrast to PASTIS, MultiSenGE provides dense labels. In this data-set, we selected 4145 SITS with a spatial dimension of \(256 \times 256\) pixels. A random split is performed to divide the data-set into training (60\%), validation (16\%) and test (24\%).  The class distribution is detailed in \cref{app:msenge-classes}. Lastly, in opposition to the two previous data-sets MultiSenGE data are pre-processed with MAJA \cite{rs11040433} instead of Sen2Cor. 
\subsubsection{CropRot Crop change detection}
\label{sec:cd}
This paper presents a new data-set for detecting abrupt changes between two SITS. Unlike the identification of changes in a time series (break detection), which may be solved by signal-based methods, the proposed task requires a more advanced semantic understanding.
More specifically, thanks to the labels provided by \emph{RPGExplorer Crop successions} \cite{levavasseur-2016-rpg-explor}, \emph{CropRot} identifies crop rotations between two consecutive years in France. In particular, the RPGExplorer database provides crop sequence labels based on the RPG (\emph{Registre Parcellaire Graphique})\footnote{\url{https://artificialisation.developpement-durable.gouv.fr/bases-donnees/registre-parcellaire-graphique}}. Within a sequence (e.g. 2015-2020), parcels are unified (each parcel has a unique identifier).

For this data-set, the following classes were selected based on the RPG labels: rapeseed, cereals, proteaginous, soybean, sunflower, maize, rice, tubers and grassland. These classes categorize vegetation based on its physiological characteristics and can be identified using RS data. Pixels that are not part of these crops for the two years 2019 and 2020 are considered as background.
Then, the label \emph{change} is assigned to pixels that have a different label between 2019 and 2020. Each data-set sample includes S2 L2A SITS for 2019 and 2020, with their corresponding labels. The label tensor has three channels containing crop labels for 2019, 2020, and change label. In our proposed downstream task, change detection is performed while ignoring background pixels.
The SITS were built using the SITS spatial extent from PASTIS where sufficient labels from the RPGExplorer were available. Due to this specific selection, the crop classes proteaginous, soybean and tuber do not appear in our data-set. Nevertheless, the code used to build this labeled data-set is published\footnote{\url{https://src.koda.cnrs.fr/iris.dumeur/modcix}}, enabling it to be extended to other regions of France and to other years. These missing classes might be integrated in an augmented version of the data-set. The change matrix between 2019 and 2020 is presented in \Cref{app:rotcrop-confmat}.
\subsection{Evaluation Protocol}
\label{sec:orgc124e6b}
\begin{figure*}[h!]
\centering
\includegraphics[width=.9\linewidth]{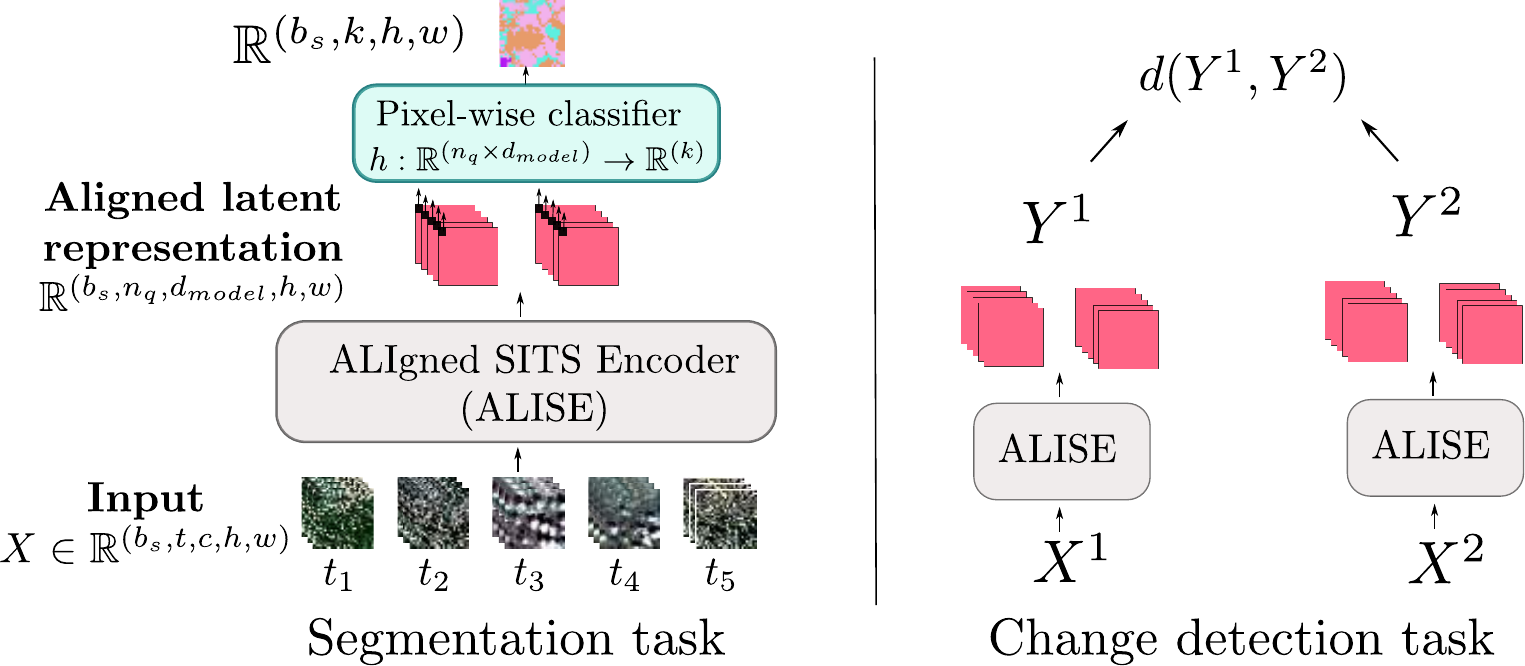}
\caption{\label{fig:downstream-tasks}The two types of downstream tasks considered. Left: segmentation task framework. A single fully-connected layer projects, for each pixel of the latent representation \(Y\), the \(n_q \times d_{model}\) features into a vector of size \(\mathbb{R}^k\) with \(k\) the number of classes. Right: change detection task between two SITS \(X^1\) and \(X^2\). The mean square error is computed between the two aligned latent representations \(Y^1\) and \(Y^2\).}
\end{figure*}

We propose two ways of exploiting ALISE representations, detailed in \autoref{fig:downstream-tasks}, corresponding to the two types of downstream tasks: semantic segmentation and change detection.
\subsubsection{Semantic segmentation tasks}
\label{sec:org18ecb47}
As detailed in \Cref{fig:downstream-tasks}, we classify each pixel-level latent vector by using a single linear layer in both segmentation tasks.  Noting the pixel-level latent vector as \(\mathbf{y}_{(b,h,w)} \in \mathbb{R}^{(d_{model}\times n_q)}\), the unnormalized logits for each class \(k\) at the pixel level can be written as: \(\mathbf{c}_{(b,h,w)}=\mathbf{y}_{(b,h,w)}A+\mathbf{b}\) where \(A \in \mathbb{R}^{(d_{model} \times n_q , k)}\) and \(\mathbf{b} \in \mathbb{R}^{k}\). The classical cross-entropy loss function is used for training\footnote{\url{https://pytorch.org/docs/stable/generated/torch.nn.CrossEntropyLoss.html}}. The latent representations are generated by a pre-trained ALISE whose weights are frozen in linear probing or updated during fine-tuning. We denote the fine-tuning and linear probing configurations as ALISE\textsuperscript{FT} and ALISE\textsuperscript{FR} respectively, while the fully supervised model is denoted as ALISE\textsuperscript{FS}. During the downstream tasks, ALISE as well as competitive models are trained with ADAM optimizer, a learning rate of 1e-4 and ReduceLROnPlateau scheduler with a patience of 10 epochs and a decay rate of 0.05.
\subsubsection{Change detection}
\label{sec:org0078d19}
As detailed in \Cref{eq:distmap}, and illustrated in \Cref{fig:downstream-tasks}, change detection between two SITS \(X^1, X^2\) is performed at a pixel level.  For each pixel located at location \((h,w)\), the mean square error between the corresponding latent vectors \(Y^1_{(\bdot,\bdot,h,w)}\) and \(Y^2_{(\bdot,\bdot,h,w)}\) is calculated. The resulting distance value serves as change detection criterion.
\begin{equation} \label{eq:distmap}
d(Y^1_{(\bdot,\bdot,h,w)},Y^2_{(\bdot,\bdot,h,w)})=\frac{1}{n_q \times d_{model}}  \sum_{n,d} (y^1_{n,d,h,w}- y^2_{n,d,h,w})^2
\end{equation}

\subsection{Competitive methods}
\label{sec:org922e1e2}
As mentioned above, ALISE is evaluated on two different types of downstream tasks: semantic segmentation and change detection. Consequently, in this section we detail the competing works associated with these two types of tasks.
\subsubsection{SITS segmentation concurrent works}
\label{sec:org0f5e86f}
ALISE can be exploited in the downstream task in  three ways : fully-supervised (FS), fine-tuned (FT) and frozen (FR).
Therefore, ALISE is compared with the following concurrent works: U-TAE \footnote{\url{https://github.com/VSainteuf/utae-paps}} \cite{garnot-2021-panop-segmen} (FS), Presto \footnote{\url{https://github.com/nasaharvest/presto}  (commit 5486fd5)\label{org3b11b71}} \cite{Tseng2023LightweightPT} (FT, FR), U-BARN \cite{dumeur-2024-self-super}(FS, FT, FR) and U-BARN-GF (FR). This last approach is a variant of U-BARN providing, as ALISE, fixed dimensional SITS representations.
\begin{figure*}[htb]
\centering
\includegraphics[width=\linewidth]{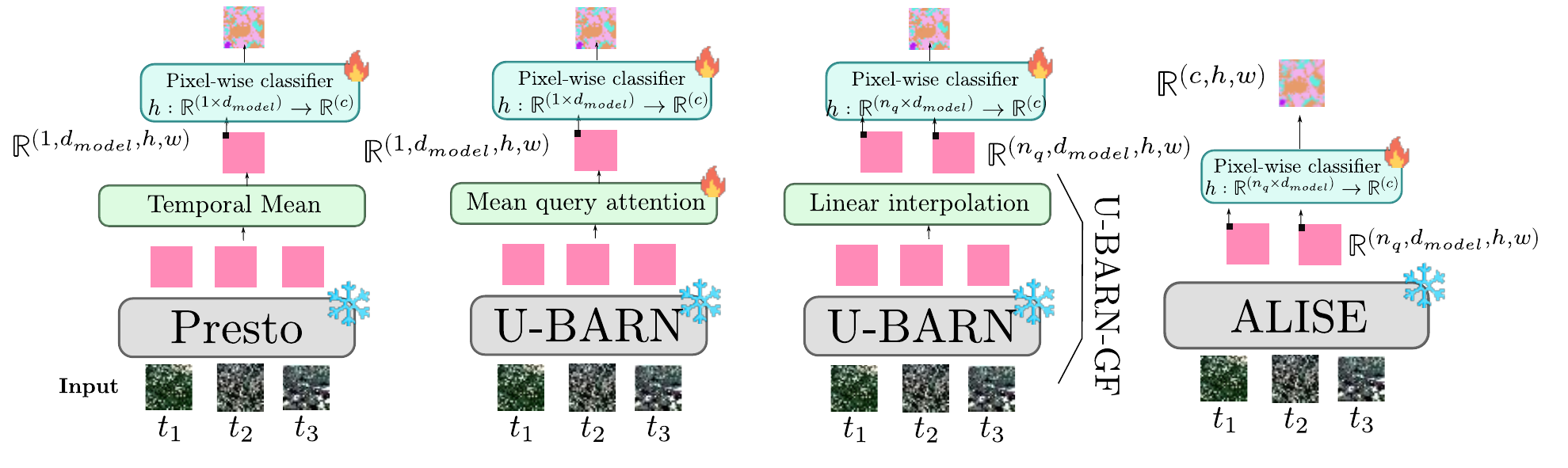}
\caption{\label{fig:baselines}Comparison of Presto, ALISE,U-BARN and U-BARN-GF when frozen for the semantic segmentation task.}
\end{figure*}
\renewcommand{\labelenumi}{(\alph{enumi})}
\begin{enumerate}
\item U-TAE is a fully supervised architecture composed of a Unet network with a lightweight temporal attention mechanism located at the bottleneck.
\item Presto is a lightweight temporal SITS encoder pre-trained as an MAE. It takes as input a monthly synthesis. The authors suggest selecting the least cloudy scene of each month. In contrast, as usually operated in RS, we train Presto with SITS composed of the median value of each band among the cloud-free acquisitions of each month. As suggested in \cite{Tseng2023LightweightPT}, to exploit the latent representations provided by Presto a temporal mean is performed.
\item U-BARN \cite{dumeur-2024-self-super} is a spatio-spectro-temporal SITS encoder pre-trained as an MAE. As U-BARN does not encode SITS into a fixed size latent representation, the shallow classifier with a mean query attention mechanism proposed in \cite{dumeur-2024-self-super} is considered here. Compared to the original implementation, we have modified the positional encoding so that U-BARN can process multi-year SITS. Besides, we have pre-trained U-BARN on MMDC-EU with the same pre-training configuration as ALISE. We call these SSL models U-BARN\textsuperscript{FT} and U-BARN\textsuperscript{FR} to denote the fine-tuning and frozen configurations.
\item U-BARN-GF. To assess the effectiveness of the proposed learnable temporal projector, we compare ALISE with an encoder composed of U-BARN followed by a linear interpolation layer, denoted U-BARN-GF. The irregular and unaligned representations from U-BARN are projected into \(n_q = 10\) regularly spaced reference dates in the temporal extent of the downstream task. The resulting aligned representations are then processed through a single fully connected layer as done with ALISE. We denote U-BARN-GF\textsuperscript{FR} the configuration where U-BARN is pre-trained and frozen during the downstream task. If the learnable projector is effective, we expect ALISE\textsuperscript{FR} to outperform U-BARN-GF\textsuperscript{FR}.
\end{enumerate}
\subsubsection{Change detection baseline}
\label{sec:org0c70107}
As indicated in \autoref{fig:downstream-tasks}, we propose to exploit ALISE representations without additional learning steps to perform change detection. To establish a fair comparison, ALISE is also compared to an unsupervised change detection strategy. In the proposed competitive work, the input SITS are interpolated onto a fixed annual common time grid using a linear interpolation (gap-filling) method. Specifically we interpolate the SITS valid acquisitions onto a regular temporal grid with a period of 5 days. The distance map is computed between the interpolated raw SITS. 

\section{Experiments}
\label{sec:results}
This section evaluates the representations provided by the pre-trained ALISE on three downstream tasks and compares them to competitive methods. First, we present a detailed analysis of ALISE’s performance in both fine-tuned and frozen configurations for the two segmentation data-sets (PASTIS and MultiSenGE). We also examine the effectiveness of the pre-training under a severe label scarcity scenario. Additionally, ALISE representations are assessed on an unsupervised change detection task with the CropRot data-set.  Next, we provide an extensive discussion on the influence of several pre-training parameters (\(t_w\), \(n_q\), \(w_{rec}\), \(w_{inv}\), \(w_{cov}\)). Lastly, we propose a qualitative assessment of the role of the learnable queries in the temporal projector.

\subsection{Segmentation tasks results}
\label{sec:org048171b}
The segmentation performances of ALISE either frozen, fine-tuned or fully-supervised on both labeled data-sets are compared here to competitive works. The two downstream data-sets differ on two main points. Firstly, in the MultiSenGE data-set, semantic labeling is dense, whereas in PASTIS, all pixels not belonging to a known crop are not classified. Consequently, we assume that spatial context must be better taken into account to succeed in the MultiSenGE task than in PASTIS. However, we assume that to distinguish between the 18 PASTIS crop classes compared to the 14 land cover classes of MultiSenGE, more complex temporal features are required. \Cref{tab:global-seg} presents the averaged F1 score, the overall accuracy (OA) and the mean intersection over Union (MIoU) on the PASTIS and MultiSenGE segmentation data-sets. For each segmentation task, detailed F1 scores per class are displayed on \autoref{tab:f1-pastis-classes} and \autoref{tab:f1-msenge-classes}, respectively. To better understand the performance differences between ALISE and Presto the confusion matrix is shown in \autoref{fig:conf-alise-presto}.

\begin{table*}[htb]
\caption{\label{tab:global-seg}F1 score averaged per class on PASTIS and MultiSenGE data-sets. The mean of the F1 scores are obtained on PASTIS’ 5 folds. On the MultiSenGE data-set, two trainings are conducted with different seeds. Each color corresponds to a pre-training configuration, and the highest score within a configuration is underlined. As no cloud masks are provided on MultiSenGE, Presto can't be assessed on this segmentation task. The number of trainable parameters are estimated on the PASTIS task.}
\centering
\begin{tabular}{|p{0.12\linewidth}|p{0.09\linewidth}|p{0.06\linewidth}|p{0.08\linewidth}|p{0.08\linewidth}|p{0.08\linewidth}|p{0.08\linewidth}|p{0.08\linewidth}|p{0.08\linewidth}|}
\hline
\textbf{Name} & \textbf{Pre-training Data-Set} & \textbf{Trainable parameters} & \textbf{PASTIS F1} & \textbf{PASTIS OA} & \textbf{PASTIS mIoU} & \textbf{MSenGE F1} & \textbf{MSenGE OA} & \textbf{MSenGE mIoU}\\
\hline
\hline
\rowcolor{green!7} ALISE\textsuperscript{FR} & MMDC-EU & 12.2K & \ul{0.68} \textpm{} 0.02 & \ul{0.85} \textpm{} 0.01 & \ul{0.56} \textpm{} 0.02 & \ul{0.17} \textpm{} 0.00 & \ul{0.57} \textpm{} 0.00 & \ul{0.11} \textpm{} 0.00\\
\rowcolor{blue!7} ALISE\textsuperscript{FT} & MMDC-EU & 1.1M & \textbf{\ul{0.81}} \textpm{} 0.02 & \textbf{\ul{0.91}} \textpm{} 0.01 & \textbf{\ul{0.70}} \textpm{} 0.02 & \textbf{\ul{0.23}} \textpm{} 0.00 & \textbf{\ul{0.63}} \textpm{} 0.00 & \textbf{\ul{0.16}} \textpm{} 0.00\\
ALISE\textsuperscript{FS} & \xmark & 1.1M & 0.80 \textpm{} 0.01 & \textbf{\ul{0.91}} \textpm{} 0.01 & 0.69 \textpm{} 0.01 & 0.21 \textpm{} 0.00 & \ul{0.62} \textpm{} 0.01 & 0.15 \textpm{} 0.00\\
\hline
\rowcolor{green!7} Presto\textsuperscript{FR} & worlwide & 2.5K & 0.27 \textpm{} 0.01 & 0.62 \textpm{} 0.00 & 0.19 \textpm{} 0.01 & \xmark & \xmark & \xmark\\
\rowcolor{blue!7} Presto\textsuperscript{FT} & worlwide & 404K & 0.55 \textpm{} 0.02 & 0.76 \textpm{} 0.01 & 0.41 \textpm{} 0.02 & \xmark & \xmark & \xmark\\
\hline
\rowcolor{green!7} U-BARN-GF\textsuperscript{FR} & MMDC-EU & 12.2K & 0.60 \textpm{} 0.01 & 0.82 \textpm{} 0.01 & 0.48 \textpm{} 0.01 & 0.12 \textpm{} 0.00 & 0.55 \textpm{} 0.00 & 0.09 \textpm{} 0.00\\
\rowcolor{green!7} U-BARN\textsuperscript{FR} & MMDC-EU & 13.8K & 0.59 \textpm{} 0.03 & 0.82 \textpm{} 0.01 & 0.47 \textpm{} 0.03 & 0.14 \textpm{} 0.00 & 0.56 \textpm{} 0.00 & 0.10 \textpm{} 0.00\\
\rowcolor{blue!7} U-BARN\textsuperscript{FT} & MMDC-EU & 1.1M & \textbf{\ul{0.81}} \textpm{} 0.02 & \textbf{\ul{0.91}} \textpm{} 0.01 & \textbf{\ul{0.70}} \textpm{} 0.02 & \textbf{\ul{0.23}} \textpm{} 0.01 & 0.62 \textpm{} 0.00 & \textbf{\ul{0.16}} \textpm{} 0.01\\
\rowcolor{black!2} U-BARN\textsuperscript{FS} & \xmark & 1.1M & 0.80 \textpm{} 0.01 & 0.90 \textpm{} 0.01 & 0.69 \textpm{} 0.02 & \textbf{\ul{0.23}} \textpm{} 0.01 & \ul{0.62} \textpm{} 0.00 & \textbf{\ul{0.16}} \textpm{} 0.01\\
\hline
\rowcolor{black!2} U-TAE & \xmark & 1.1M & \textbf{\ul{0.81}} \textpm{} 0.02 & \textbf{\ul{0.91}} \textpm{} 0.01 & \textbf{\ul{0.70}} \textpm{} 0.03 & 0.15 \textpm{} 0.02 & 0.61 \textpm{} 0.01 & 0.11 \textpm{} 0.01\\
\hline
\end{tabular}
\end{table*}
First, although this paper does not focus on the construction of a novel fully supervised framework for SITS, ALISE\textsuperscript{FS} architecture achieves performances consistent with current SOTA (U-TAE, U-BARN\textsuperscript{FS}).
Next, \autoref{tab:global-seg} demonstrates that ALISE\textsuperscript{FR} outperforms the existing models such as Presto\textsuperscript{FR} and U-BARN\textsuperscript{FR} on the PASTIS data-set.
Besides, compared to the PASTIS segmentation task, performances are lower on MultiSenGE. This may be explained by the fact that this last data-set is highly imbalanced, and minority classes decrease the macro-averaged mIoU and F1 scores as shown in \autoref{tab:f1-msenge-classes}. Furthermore, the fine-tuned configuration (ALISE\textsuperscript{FT}) does not significantly outperform the fully-supervised approach (ALISE\textsuperscript{FS}). This finding is nonetheless consistent with the previous study \cite{dumeur-2024-self-super} conducted on U-BARN.
Differences between ALISE and the other two competitive works are illustrated in \autoref{fig:baselines} and further detailed below.
\subsubsection{ALISE vs U-BARN}
\label{sec:orgf8510f7}
Segmentation metrics detailed  in \autoref{tab:global-seg}, show that ALISE\textsuperscript{FR} outperforms U-BARN\textsuperscript{FR}:  (+9\% F1), (+3\% 0A) and  (+9\% mIoU)  on the PASTIS data-set and (+3\% F1), (+1\%OA) and (+1\% on mIoU) on MultiSenGE.  Remarkably, ALISE significantly outperforms U-BARN in linear probing, while having a shallower classifier (no learnable mean query) and smaller latent representations (10 latent temporal features). We also observe on \autoref{tab:f1-pastis-classes}, that the boost of performance may vary depending on the PASTIS crop classes. Several classes such as winter tricitale (+25\%), sunflower (+14\%), sorghum (+19\%) and mixed cereal (+15\%) exhibit stronger boost of performances compared to the other classes. 
The overall gain of performance can be explained by the differences between ALISE and U-BARN. ALISE differs from U-BARN in two main aspects: (i) its encoder provides fixed-size, aligned representations, and (ii) the pre-training strategy is different. As detailed in \autoref{sec:alise},  ALISE corresponds to the U-BARN architecture on top of which is placed a temporal projector. Experiments detailed in \autoref{sec:ablation-study} show that ALISE’s pre-training is primarily driven by its cross-reconstruction task, which is close to U-BARN’s MAE pre-training. Therefore, we believe that the improvement in performance when freezing the pre-trained SITS encoder is largely due to the inclusion of the temporal projector in ALISE. 
Furthermore, to ensure that these improved results are not caused by the use of the shallow classifier architecture combined with ALISE, ALISE\textsuperscript{FR} performances are compared with U-BARN-GF.  Similarly to ALISE, U-BARN-GF aligned features are injected into a  fully-connected layer which operate on the spectro-spatio-temporal features. We observe in \autoref{tab:global-seg} that U-BARN\textsuperscript{FR} and U-BARN-GF\textsuperscript{FR} have close performances and that ALISE\textsuperscript{FR} outperforms U-BARN-GF\textsuperscript{FR} on both data-sets. This result demonstrates the effectiveness of using a learnable temporal projector over a linear interpolation strategy. Furthermore, by design, ALISE provides more relevant representations in the frozen configuration than U-BARN.
\subsubsection{ALISE vs Presto}
\label{sec:orgcc890ba}
\begin{figure*}[htb]
\centering
\includegraphics[width=0.8\linewidth]{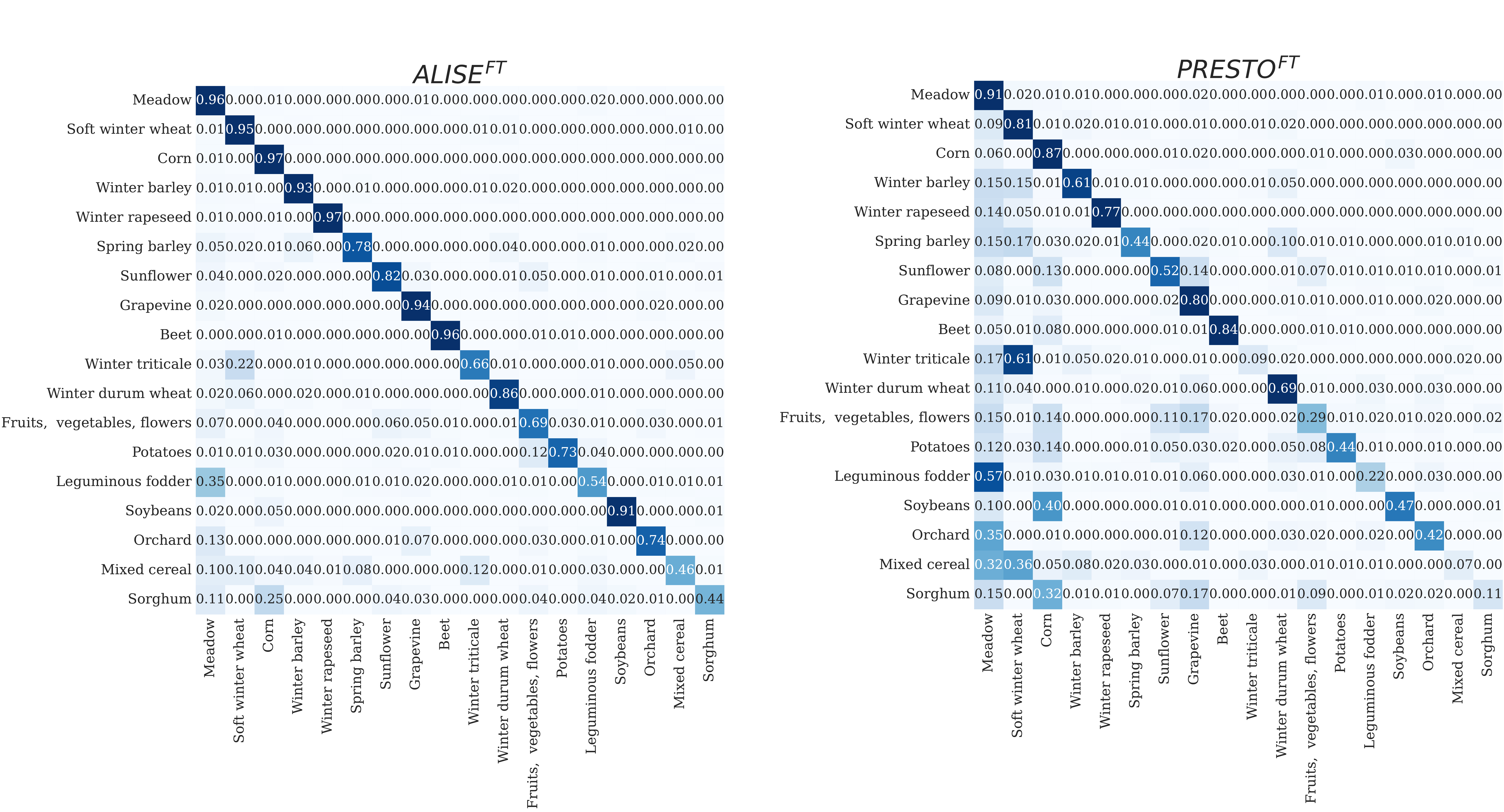}
\caption{\label{fig:conf-alise-presto}Confusion matrices on the PASTIS crop segmentation data-set obtained after fine-tuning. For each confusion matrix, rows correspond to true labels and columns to predictions. The matrices are normalized per row. On the left obtained with ALISE\textsuperscript{FT} and on the right Presto\textsuperscript{FT}.}
\end{figure*}

We observe that ALISE\textsuperscript{FR} outperforms both frozen and fine-tuned Presto configurations, by 41.5\% and 13,6\%, respectively. To study more deeply the obtained performances, we compare the confusion matrices obtained by ALISE\textsuperscript{FT} and Presto\textsuperscript{FT} in \autoref{fig:conf-alise-presto}. We observe that ALISE and Presto show similar confusions between classes: winter triticale and soft winter wheat, leguminous fodder and meadow, corn and sorghum. Nevertheless, the confusion values between classes obtained by ALISE are each time lower than values reached by Presto. We posit that this result can be explained by several factors. Firstly, Presto is a lightweight spectro-temporal architecture that does not take spatial context into account. Such a design may not be relevant to segmentation tasks. Additionally, due to the required under-sampling protocol (Presto exploits monthly synthesis instead of all available acquisitions), it may miss key temporal information in comparison to ALISE. Furthermore, the implemented temporal positional encoding in the released code\textsuperscript{\ref{org3b11b71}} raises questions. Traditionally, in the classical transformer model, the positional encoding is added or concatenated to the input data along the channel dimension. However, from our understanding of the code, in the proposed implementation, the positional encoding is concatenated along the temporal dimension. 

\begin{table*}[htbp]
\caption{\label{tab:f1-pastis-classes}F1 score per class on PASTIS dataset for each training configuration. Mean and standard deviation of the F1 score obtained using k-fold training are detailed.}
\centering
\begin{tabular}{|p{0.13\linewidth}|p{0.075\linewidth}|p{0.075\linewidth}|p{0.075\linewidth}|p{0.075\linewidth}|p{0.08\linewidth}|p{0.08\linewidth}|p{0.08\linewidth}|p{0.08\linewidth}|p{0.075\linewidth}|}
\hline
 & \textbf{ALISE\textsuperscript{FR}} & \textbf{ALISE\textsuperscript{FT}} & \textbf{ALISE\textsuperscript{FS}} & \textbf{Presto\textsuperscript{FR}} & \textbf{Presto\textsuperscript{FT}} & \textbf{U-BARN\textsuperscript{FR}} & \textbf{U-BARN\textsuperscript{FT}} & \textbf{U-BARN\textsuperscript{FS}} & \textbf{U-TAE}\\
\hline
\rowcolor{black!10} Meadow & 0.91 \textpm{} 0.01 & \textbf{0.94} \textpm{} 0.01 & \textbf{0.94} \textpm{} 0.01 & 0.75 \textpm{} 0.01 & 0.85 \textpm{} 0.01 & 0.90 \textpm{} 0.01 & \textbf{0.94} \textpm{} 0.01 & \textbf{0.94} \textpm{} 0.01 & \textbf{0.94} \textpm{} 0.01\\
Soft winter wheat & 0.89 \textpm{} 0.01 & \textbf{0.94} \textpm{} 0.01 & \textbf{0.94} \textpm{} 0.01 & 0.65 \textpm{} 0.03 & 0.79 \textpm{} 0.02 & 0.86 \textpm{} 0.01 & \textbf{0.94} \textpm{} 0.01 & \textbf{0.94} \textpm{} 0.01 & \textbf{0.94} \textpm{} 0.01\\
\rowcolor{black!10} Corn & 0.93 \textpm{} 0.01 & \textbf{0.96} \textpm{} 0.01 & \textbf{0.96} \textpm{} 0.00 & 0.71 \textpm{} 0.01 & 0.85 \textpm{} 0.01 & 0.91 \textpm{} 0.01 & \textbf{0.96} \textpm{} 0.01 & \textbf{0.96} \textpm{} 0.01 & 0.96 \textpm{} 0.01\\
Winter barley & 0.82 \textpm{} 0.02 & \textbf{0.92} \textpm{} 0.01 & \textbf{0.92} \textpm{} 0.02 & 0.21 \textpm{} 0.04 & 0.67 \textpm{} 0.03 & 0.77 \textpm{} 0.04 & \textbf{0.92} \textpm{} 0.02 & \textbf{0.92} \textpm{} 0.02 & \textbf{0.92} \textpm{} 0.01\\
\rowcolor{black!10} Winter rapeseed & 0.91 \textpm{} 0.01 & \textbf{0.96} \textpm{} 0.01 & \textbf{0.96} \textpm{} 0.01 & 0.49 \textpm{} 0.04 & 0.81 \textpm{} 0.02 & 0.87 \textpm{} 0.04 & \textbf{0.96} \textpm{} 0.01 & \textbf{0.96} \textpm{} 0.01 & \textbf{0.96} \textpm{} 0.01\\
Spring barley & 0.67 \textpm{} 0.08 & \textbf{0.79} \textpm{} 0.05 & 0.77 \textpm{} 0.06 & 0.01 \textpm{} 0.01 & 0.49 \textpm{} 0.07 & 0.61 \textpm{} 0.09 & 0.78 \textpm{} 0.06 & 0.76 \textpm{} 0.06 & 0.77 \textpm{} 0.05\\
\rowcolor{black!10} Sunflower & 0.71 \textpm{} 0.03 & \textbf{0.83} \textpm{} 0.01 & \textbf{0.83} \textpm{} 0.02 & 0.18 \textpm{} 0.03 & 0.56 \textpm{} 0.05 & 0.57 \textpm{} 0.07 & \textbf{0.83} \textpm{} 0.04 & 0.81 \textpm{} 0.03 & \textbf{0.83} \textpm{} 0.05\\
Grapevine & 0.82 \textpm{} 0.02 & \textbf{0.92} \textpm{} 0.01 & 0.91 \textpm{} 0.01 & 0.57 \textpm{} 0.06 & 0.71 \textpm{} 0.05 & 0.77 \textpm{} 0.03 & 0.91 \textpm{} 0.01 & 0.91 \textpm{} 0.00 & 0.91 \textpm{} 0.01\\
\rowcolor{black!10} Beet & 0.93 \textpm{} 0.02 & \textbf{0.96} \textpm{} 0.01 & 0.95 \textpm{} 0.02 & 0.45 \textpm{} 0.10 & 0.85 \textpm{} 0.02 & 0.89 \textpm{} 0.01 & \textbf{0.96} \textpm{} 0.01 & \textbf{0.96} \textpm{} 0.02 & \textbf{0.96} \textpm{} 0.01\\
Winter triticale & 0.41 \textpm{} 0.06 & 0.70 \textpm{} 0.06 & 0.66 \textpm{} 0.06 & 0.00 \textpm{} 0.00 & 0.15 \textpm{} 0.05 & 0.16 \textpm{} 0.06 & 0.69 \textpm{} 0.04 & 0.65 \textpm{} 0.07 & \textbf{0.73} \textpm{} 0.04\\
\rowcolor{black!10} Winter durum wheat & 0.74 \textpm{} 0.03 & \textbf{0.83} \textpm{} 0.03 & 0.81 \textpm{} 0.03 & 0.41 \textpm{} 0.02 & 0.64 \textpm{} 0.03 & 0.69 \textpm{} 0.01 & 0.82 \textpm{} 0.03 & 0.80 \textpm{} 0.02 & 0.82 \textpm{} 0.04\\
Fruits/veg/flowers & 0.49 \textpm{} 0.07 & 0.70 \textpm{} 0.03 & 0.65 \textpm{} 0.03 & 0.04 \textpm{} 0.02 & 0.34 \textpm{} 0.10 & 0.35 \textpm{} 0.05 & \textbf{0.71} \textpm{} 0.03 & 0.65 \textpm{} 0.03 & 0.69 \textpm{} 0.06\\
\rowcolor{black!10} Potatoes & 0.65 \textpm{} 0.07 & \textbf{0.76} \textpm{} 0.08 & 0.70 \textpm{} 0.05 & 0.10 \textpm{} 0.06 & 0.52 \textpm{} 0.09 & 0.55 \textpm{} 0.09 & 0.74 \textpm{} 0.03 & 0.70 \textpm{} 0.06 & 0.71 \textpm{} 0.09\\
Leguminous fodder & 0.44 \textpm{} 0.08 & 0.60 \textpm{} 0.07 & \textbf{0.64} \textpm{} 0.07 & 0.10 \textpm{} 0.02 & 0.30 \textpm{} 0.05 & 0.35 \textpm{} 0.11 & 0.63 \textpm{} 0.07 & 0.63 \textpm{} 0.09 & 0.61 \textpm{} 0.08\\
\rowcolor{black!10} Soybeans & 0.82 \textpm{} 0.04 & 0.92 \textpm{} 0.02 & \textbf{0.93} \textpm{} 0.02 & 0.01 \textpm{} 0.02 & 0.54 \textpm{} 0.06 & 0.71 \textpm{} 0.07 & \textbf{0.93} \textpm{} 0.02 & 0.92 \textpm{} 0.02 & 0.92 \textpm{} 0.03\\
Orchard & 0.61 \textpm{} 0.05 & 0.77 \textpm{} 0.04 & 0.77 \textpm{} 0.04 & 0.12 \textpm{} 0.04 & 0.47 \textpm{} 0.07 & 0.55 \textpm{} 0.04 & 0.77 \textpm{} 0.05 & 0.77 \textpm{} 0.04 & \textbf{0.79} \textpm{} 0.04\\
\rowcolor{black!10} Mixed cereal & 0.23 \textpm{} 0.08 & 0.53 \textpm{} 0.07 & 0.53 \textpm{} 0.05 & 0.00 \textpm{} 0.00 & 0.11 \textpm{} 0.04 & 0.08 \textpm{} 0.05 & 0.54 \textpm{} 0.07 & 0.54 \textpm{} 0.05 & \textbf{0.55} \textpm{} 0.06\\
Sorghum & 0.29 \textpm{} 0.08 & 0.50 \textpm{} 0.11 & 0.50 \textpm{} 0.09 & 0.00 \textpm{} 0.00 & 0.18 \textpm{} 0.08 & 0.10 \textpm{} 0.09 & 0.52 \textpm{} 0.11 & 0.49 \textpm{} 0.12 & \textbf{0.53} \textpm{} 0.13\\
\hline
\end{tabular}
\end{table*}

\begin{table*}[htbp]
\caption{\label{tab:f1-msenge-classes}F1 score per class on the MultiSenGE segmentation task for each configuration. Two trainings are performed for each configuration.}
\centering
\begin{tabular}{|p{0.23\linewidth}|p{0.075\linewidth}|p{0.075\linewidth}|p{0.075\linewidth}|p{0.075\linewidth}|p{0.08\linewidth}|p{0.08\linewidth}|p{0.08\linewidth}|}
\hline
 & \textbf{ALISE\textsuperscript{FS}} & \textbf{ALISE\textsuperscript{FT}} & \textbf{ALISE\textsuperscript{FR}} & \textbf{U-BARN\textsuperscript{FR}} & \textbf{U-BARN\textsuperscript{FS}} & \textbf{U-BARN\textsuperscript{FT}} & \textbf{U-TAE}\\
\hline
\rowcolor{black!10} Dense Built-Up & \textbf{0.04} \textpm{} 0.00 & 0.00 \textpm{} 0.00 & 0.00 \textpm{} 0.00 & 0.00 \textpm{} 0.00 & 0.00 \textpm{} 0.00 & 0.00 \textpm{} 0.00 & 0.00 \textpm{} 0.00\\
Sparse Built-Up & 0.11 \textpm{} 0.02 & 0.13 \textpm{} 0.00 & 0.02 \textpm{} 0.01 & 0.01 \textpm{} 0.00 & 0.14 \textpm{} 0.02 & 0.14 \textpm{} 0.02 & \textbf{0.15} \textpm{} 0.01\\
\rowcolor{black!10} Specialized Built-Up Areas & 0.13 \textpm{} 0.00 & \textbf{0.31} \textpm{} 0.01 & 0.07 \textpm{} 0.00 & 0.02 \textpm{} 0.00 & 0.27 \textpm{} 0.03 & 0.27 \textpm{} 0.09 & 0.02 \textpm{} 0.02\\
Specialized but Vegetative Areas & 0.00 \textpm{} 0.00 & 0.00 \textpm{} 0.00 & 0.00 \textpm{} 0.00 & 0.00 \textpm{} 0.00 & 0.00 \textpm{} 0.00 & 0.00 \textpm{} 0.00 & 0.00 \textpm{} 0.00\\
\rowcolor{black!10} Large Scale Networks & 0.00 \textpm{} 0.00 & 0.00 \textpm{} 0.00 & 0.00 \textpm{} 0.00 & 0.00 \textpm{} 0.00 & 0.00 \textpm{} 0.00 & 0.00 \textpm{} 0.00 & 0.00 \textpm{} 0.00\\
Arable Lands & 0.67 \textpm{} 0.01 & \textbf{0.68} \textpm{} 0.00 & 0.62 \textpm{} 0.00 & 0.62 \textpm{} 0.00 & \textbf{0.68} \textpm{} 0.00 & \textbf{0.68} \textpm{} 0.00 & 0.67 \textpm{} 0.01\\
\rowcolor{black!10} Vineyards & 0.46 \textpm{} 0.04 & \textbf{0.51} \textpm{} 0.01 & 0.32 \textpm{} 0.00 & 0.21 \textpm{} 0.04 & 0.47 \textpm{} 0.04 & 0.48 \textpm{} 0.02 & 0.19 \textpm{} 0.27\\
Orchards & 0.00 \textpm{} 0.00 & 0.00 \textpm{} 0.00 & 0.00 \textpm{} 0.00 & 0.00 \textpm{} 0.00 & 0.00 \textpm{} 0.00 & 0.00 \textpm{} 0.00 & 0.00 \textpm{} 0.00\\
\rowcolor{black!10} Grasslands & 0.39 \textpm{} 0.04 & 0.39 \textpm{} 0.00 & 0.33 \textpm{} 0.01 & 0.31 \textpm{} 0.02 & 0.40 \textpm{} 0.01 & \textbf{0.41} \textpm{} 0.01 & 0.40 \textpm{} 0.00\\
Groces,Hegdes & 0.00 \textpm{} 0.00 & 0.00 \textpm{} 0.00 & 0.00 \textpm{} 0.00 & 0.00 \textpm{} 0.00 & 0.00 \textpm{} 0.00 & 0.00 \textpm{} 0.00 & 0.00 \textpm{} 0.00\\
\rowcolor{black!10} Forest & 0.72 \textpm{} 0.01 & \textbf{0.73} \textpm{} 0.00 & 0.67 \textpm{} 0.00 & 0.66 \textpm{} 0.00 & \textbf{0.73} \textpm{} 0.00 & \textbf{0.73} \textpm{} 0.00 & 0.71 \textpm{} 0.02\\
Open Spaces,Mineral & 0.00 \textpm{} 0.00 & 0.00 \textpm{} 0.00 & 0.00 \textpm{} 0.00 & 0.00 \textpm{} 0.00 & 0.00 \textpm{} 0.00 & 0.00 \textpm{} 0.00 & 0.00 \textpm{} 0.00\\
\rowcolor{black!10} Wetlands & 0.00 \textpm{} 0.00 & 0.00 \textpm{} 0.00 & 0.00 \textpm{} 0.00 & 0.00 \textpm{} 0.00 & 0.00 \textpm{} 0.00 & 0.00 \textpm{} 0.00 & 0.00 \textpm{} 0.00\\
\hline
\end{tabular}
\end{table*}

\subsection{Label scarcity scenario}
\label{sec:orgb1b4df9}
To assess the behavior of the model in a severe label scarcity scenario, a reduced version of the PASTIS data-set has been created. Following the approach of \cite{dumeur-2024-self-super}, we have used five smaller data-sets, each composed of 30 SITS for each PASTIS fold.
 Therefore, the results shown in \autoref{tab:expe-scarcity} correspond to the averaged macro F1 score across 25 trials.
Under severe label scarcity, the fine-tuned model outperforms the fully-supervised framework by 12.5\%. Interestingly, the frozen ALISE also outperforms its fully-supervised configuration by 9.7\%. Given its reduced number of pre-trainable parameters compared to fully-supervised and fine-tuned approaches, ALISE\textsuperscript{FR} is an ideal candidate for scenarios with limited labeled data.
\begin{table}[htbp]
\caption{\label{tab:expe-scarcity}Macro-averaged F1 scores obtained on PASTIS on labeled data scarcity scenario. Each PASTIS fold is composed of 30 labeled SITS.}
\centering
\begin{tabular}{ll}
Model & F1\\
\hline
\rowcolor{green!7} ALISE\textsuperscript{FR} & 0.44 \textpm{} 0.01\\
\rowcolor{blue!7} ALISE\textsuperscript{FT} & \textbf{0.47} \textpm{} 0.04\\
ALISE\textsuperscript{FS} & 0.34 \textpm{} 0.06\\
\end{tabular}
\end{table}

\subsection{Change detection task}
\label{sec:org3ac7f5c}
\begin{figure*}[htb]
\centering
\includegraphics[width=.9\linewidth]{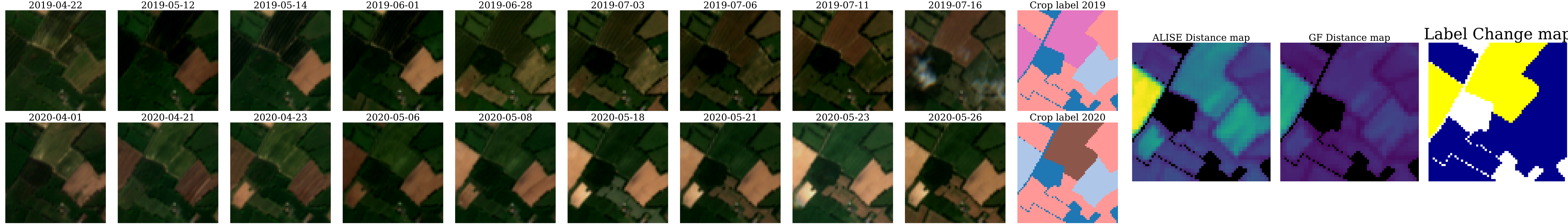}
\caption{\label{fig:ex-cd}Visualization of a change map obtained on the change detection data-set with the pre-trained ALISE. The top and bottom rows represent a portion of the S2 SITS along with their crop classes for 2019 and 2020, respectively. These SITS portions have similar index position within their SITS. In the crop label maps, dark blue represents the background class. To the right, the distance maps computed from the aligned representations from ALISE and the Gap-Filling methods are shown. The same scale is used in the colorbar of the distance maps. Pixels that belong to the background class are masked. On the far right, the label change map is represented with, in white the background class, in blue the \emph{no change} label, and in yellow the \emph{change} label.}
\end{figure*}

To quantitatively compare the change detection performances on the proposed \emph{CropRot} dataset, we compute the area under the receiver operating characteristic curve (AUC) score. This score is calculated on the distance map computed between the representations of SITS from two different years.   \Cref{tab:cd} shows that better change detection results are obtained by comparing SITS representations encoded by ALISE. Besides, in contrast to the linear interpolation on the raw input SITS, ALISE change detection framework does not require cloud mask information. Furthermore, our results demonstrate that ALISE’s representations are relevant for change detection, even though its positional encoding information is absolute and not relative to a year (day of the year). In other words, ALISE can still learn SITS invariance between different years, even if the positional encoding differs for each of the two years being compared. 
\begin{table}[t!]
\caption{\label{tab:cd}Area under the ROC curve metric on CropRot.}
\centering
\begin{tabular}{ll}
\hline
Input Data & AUC\\
\hline
\rowcolor{black!10} ALISE representations & \textbf{0.91}\\
raw interpolated SITS & 0.88\\
\hline
\end{tabular}
\end{table}
In addition, a qualitative analysis of the change detection maps is performed. Given two annual irregular and unaligned SITS from 2019 and 2020, \Cref{fig:ex-cd} illustrates the obtained change maps. In this example SITS, the number of available spring acquisitions is greater in 2020 than in 2019.
Besides, \Cref{fig:ex-cd} shows class variability even when there is no change. These variations can be caused by different agricultural practices, meteorological events, and different acquisition dates. In \autoref{fig:ex-cd}, this intra-class variability is observed when looking at the fields located at the center bottom of the SITS (red circle). Although the crop class of this field has not changed between 2019 and 2020, we can visually observe important differences between 14/05/2019 and 18/05/2020 which are supposed to be close acquisitions. Nevertheless, the distance map shown in \Cref{fig:ex-cd} is not affected by such input intra-class variability. In addition, compared to the distance map obtained from interpolated raw SITS, the distance on ALISE representations better distinguishes modified crops from unchanged ones. 
\subsection{Co-influence of \(t_w, w_{inv}, w_{cov}, w_{rec}\)}
\label{sec:ablation-study}
\(t_w\) is a hyper-parameter involved in the view generation process and detailed in \autoref{sec:view-gene}. More precisely, \(t_w\) is the number of acquisitions contained in the temporal window used to build each view. Increasing \(t_w\) is supposed to create greater discrepancies between views therefore impacting both the discriminative and the cross-reconstruction losses. As \(t_w\) is the number of consecutive acquisitions, when \(t_w\) is increased, the cross-reconstruction task is no longer a simple interpolation task.
Therefore, we aim to assess the co-influence of the view generation protocol (controlled by \(t_w\)) and the losses weights. Different pre-training configurations evaluating the impact of the four parameters \((t_w, w_{inv}, w_{cov}, w_{rec})\)  have been performed. For each configuration, results obtained on the five PASTIS folds are averaged by considering four different pre-trained models with different seeds. Only the loss weights and the \(t_w\) parameter are evaluated, all the rest of hyper-parameters are fixed for the rest of pre-trained model configurations. The covariance weight value is set to 0.05 to reproduce the balance between the invariance and covariance losses suggested in VicReg \cite{bardes2022vicreg}. The influence of \(t_w\) is analyzed by studying the macro averaged F1 score before providing a more precise analysis per crop class.
\subsubsection{Macro averaged F1 score}
\label{sec:org614f40f}
\begin{figure}[t!]
\centering
\includegraphics[width=.9\linewidth]{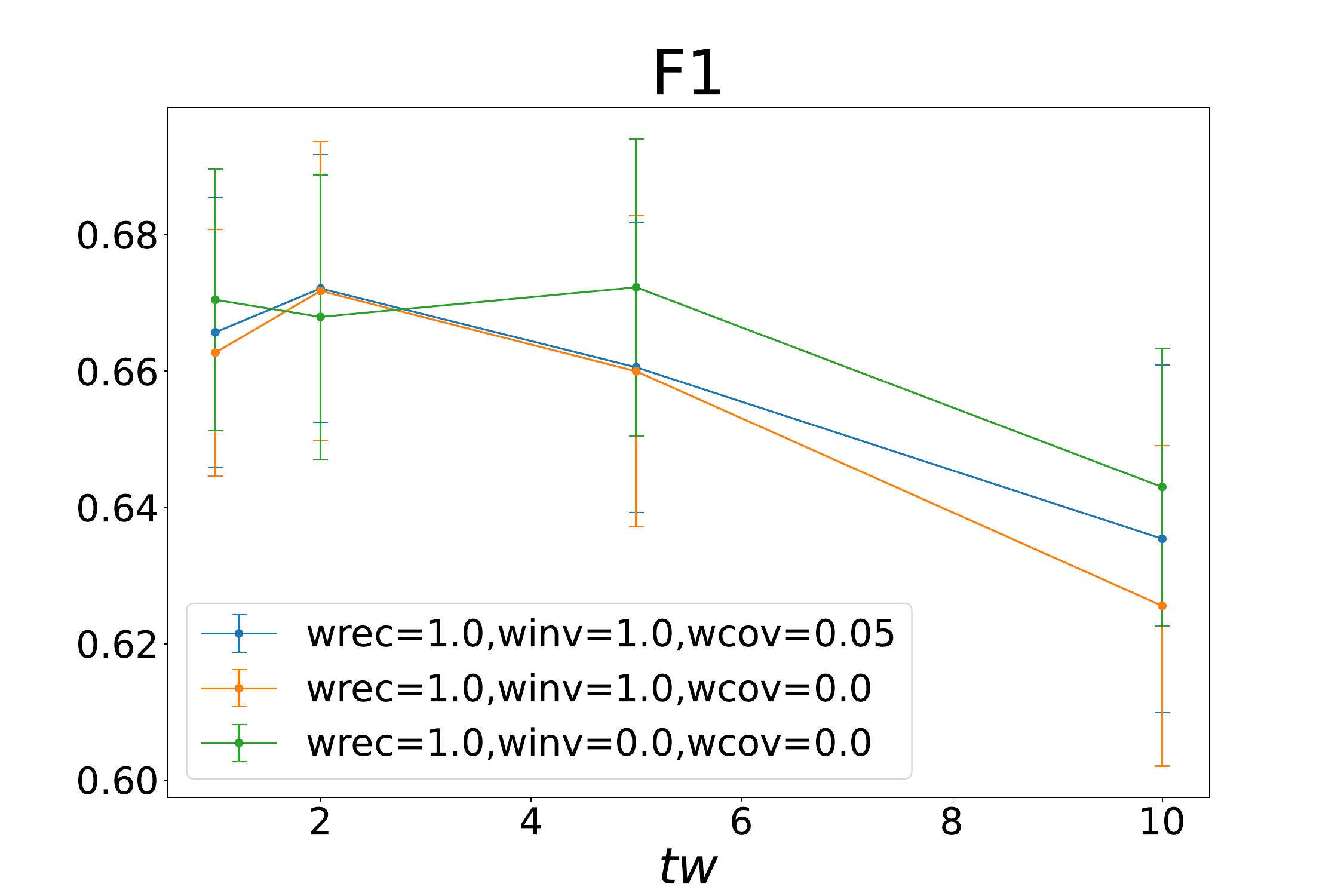}
\caption{\label{fig:influ-tw}Segmentation task performances on PASTIS linear probing as a function of \(t_w\). In all these experiments \(n_q=10\), 4 pre-trainings were conducted and their performances on 5 of PASTIS folds were evaluated.}
\end{figure}
 \Cref{fig:influ-tw} displays the linear probing performances as a function of \(t_w\). This figure shows that the additional invariance latent loss (\(L_{inv}\)) significantly degrades the linear probing performances  for \(t_w\) greater than 2,  (the orange and blue curves are lower than the green one ). We assume that the invariance loss might constrain too much the total loss when pre-training data is composed of large dissimilar views. For \(t_w=2\), there seems to be a slight improvement in the linear probing performances when invariance loss is incorporated. Lastly, these experiments do not show any benefit from using the covariance loss (\(L_{cov}\)) in addition to the invariance loss. There are several possible explanations for this outcome. First, the large memory size of SITS limits the batch size, therefore experiments have been conducted with a batch size equal to 2. Although we use \(b \times n_q \times h \times w\) samples to estimate the covariance matrix, these samples are correlated. In the original VicReg implementation \cite{bardes2022vicreg}, the covariance was estimated across 2048 samples, each corresponding to a different image. Second, the covariance loss in VicReg plays a crucial role in preventing information collapse. In our framework, the cross-reconstruction loss prevents collapsing, making the covariance loss less important during pre-training. Third, the projector architecture may impact the computation of the covariance loss. Therefore, more experiments studying the impact of the batch size and the projector could be necessary.
Lastly, the green curve in \autoref{fig:influ-tw} depicts the influence of \(t_w\) when solely the reconstruction loss is applied. In this case, the downstream segmentation performance is impacted also by \(t_w\). With large temporal windows (\(t_w=10\)), the pre-training reconstruction task may become too complex, which prevents the model from learning informative SITS representations. Surprisingly, with smaller values of \(t_w \le 5\), no major differences are observed. This could be explained by the fact that, unlike regular time series processing, \(t_w\) does not control the temporal extent that is reconstructed. Inherent important temporal gaps in S2 SITS  might provide a complex pre-training task even with \(t_w=1\). 
\subsubsection{F1 score per class}
\label{sec:orgf36e266}
\begin{figure*}[h!]
\centering
\includegraphics[width=.9\linewidth]{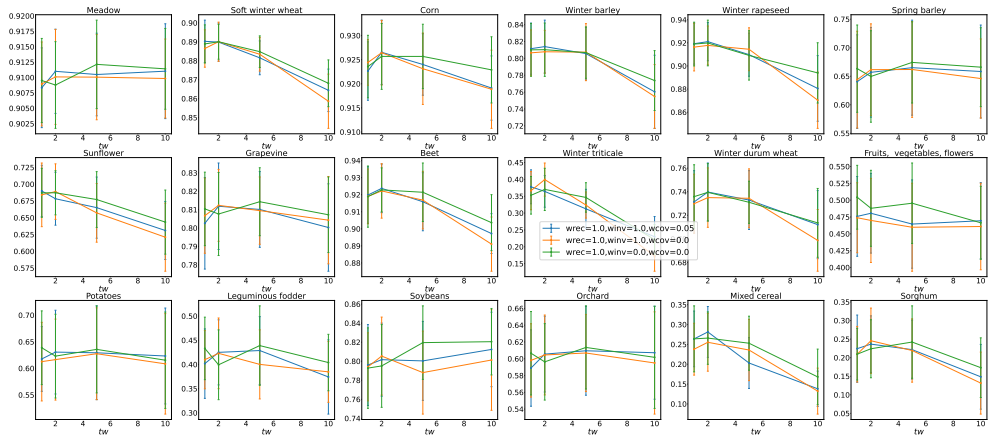}
\caption{\label{fig:influ-tw-classes}F1 score per class on PASTIS linear probing as a function of \(t_w\). In all these experiments \(n_q=10\). Results are averaged over 4 pre-trained models for all 5 PASTIS folds.}
\end{figure*}

We propose a more in-depth analysis of the effect of \(t_w\) and the pre-training loss weights in \autoref{fig:influ-tw-classes}. Notably, similar to the previous experiment, the F1 score for each PASTIS crop class is plotted as a function of \(t_w\). Different behaviors are observed depending on the crop classes. For many crop classes, there is a decrease in the F1 score when \(t_w\) increases. However, some crop classes such as meadow, corn, spring barley, grapevine, fruits, vegetables \& flowers, potatoes, leguminous fodder, and orchard are unaffected by \(t_w\).
 With the exception of grassland, maize and spring barley, we hypothesize that the lack of \(t_w\) effect for these classes may be linked to the fact that they are either permanent (grapevine, orchards) or greenhouse. Interestingly, the soybean class exhibits an outlier behavior, with an increase in F1 score as \(t_w\) increases. This experiment demonstrates that the influence of pre-training conditions differs depending on the target class.
\subsection{Impact of \(n_q\)}
\label{sec:org4a622ec}
\begin{figure}[h!]
\centering
\includegraphics[width=0.6\linewidth]{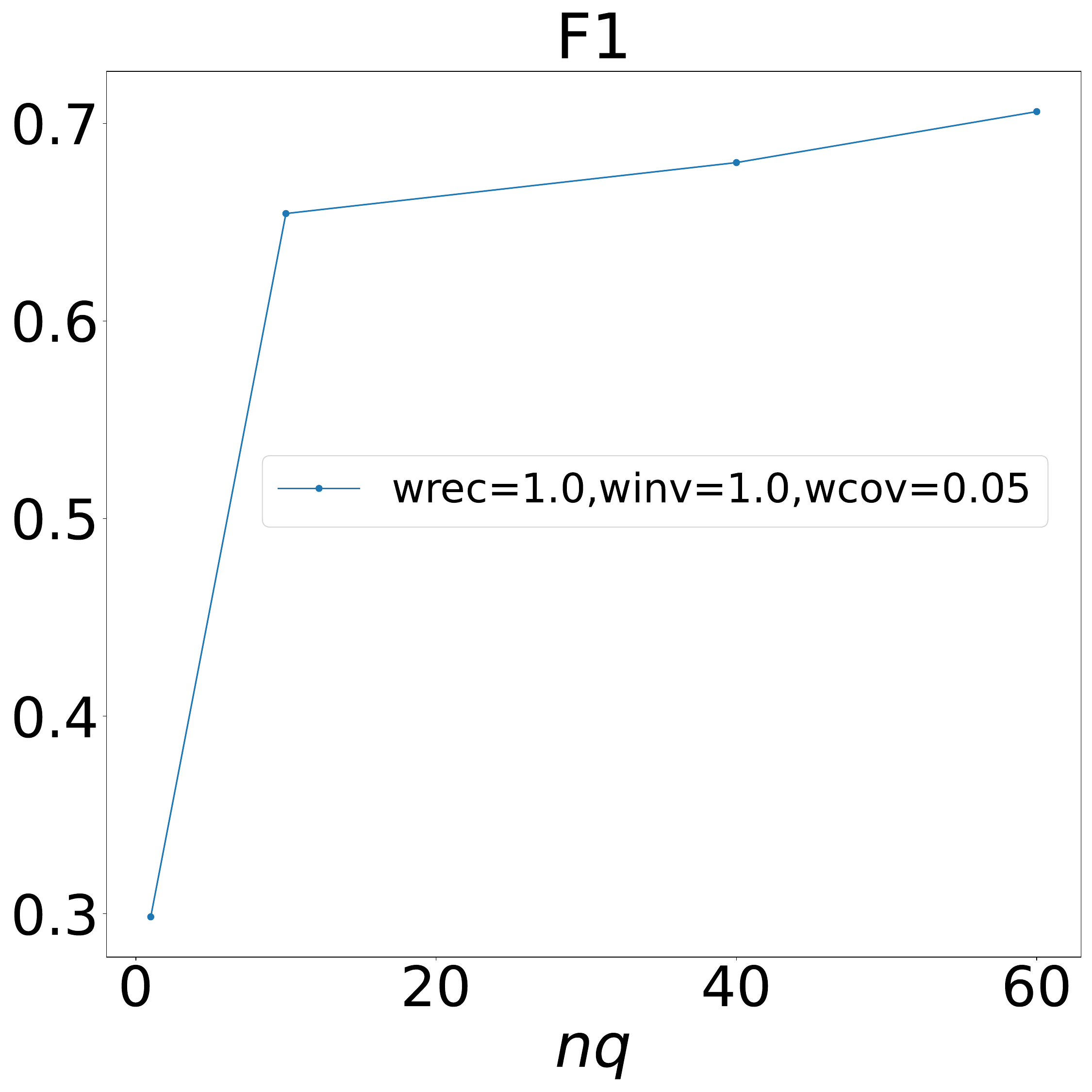}
\caption{\label{fig:influ-nq}Segmentation task performances in linear probing configuration on the PASTIS data-set as a function of \(n_q\). In these experiments, \(t_w = 5\). For each configuration, one pre-training is conducted and the downstream task is performed on one out of five PASTIS experiments.}
\end{figure}

For practical purposes, it is relevant to reduce the size of the latent representation (\(n_q\)) while preserving the downstream tasks performances. \Cref{fig:influ-nq} plots the segmentation performances on the PASTIS data-set as a function of \(n_q\). For each configuration, one pre-training was conducted, and the performances were assessed on one out of the five available PASTIS experiments. 
We observe that increasing the value of \(n_q\) improves downstream task F1 score. This can be attributed to two factors. Firstly, a greater value of \(n_q\) means a larger latent space, and therefore potentially more information contained within it. Secondly, it is assumed that a smaller value of \(n_q\) makes the cross-reconstruction task more difficult due to the compression performed in the proposed temporal projector. This compression could penalize the cross-reconstruction pre-training task. Nevertheless, these comparisons with different possible values of \(n_q\) may not be totally fair. Indeed, a higher value of \(n_q\) results in a larger classifier during linear probing, which may have a positive impact on downstream task performances. 
\subsection{Qualitative analysis of the temporal projector}
\label{sec:org47ba9fb}
\begin{figure*}[htb]
\centering
\includegraphics[width=.9\linewidth]{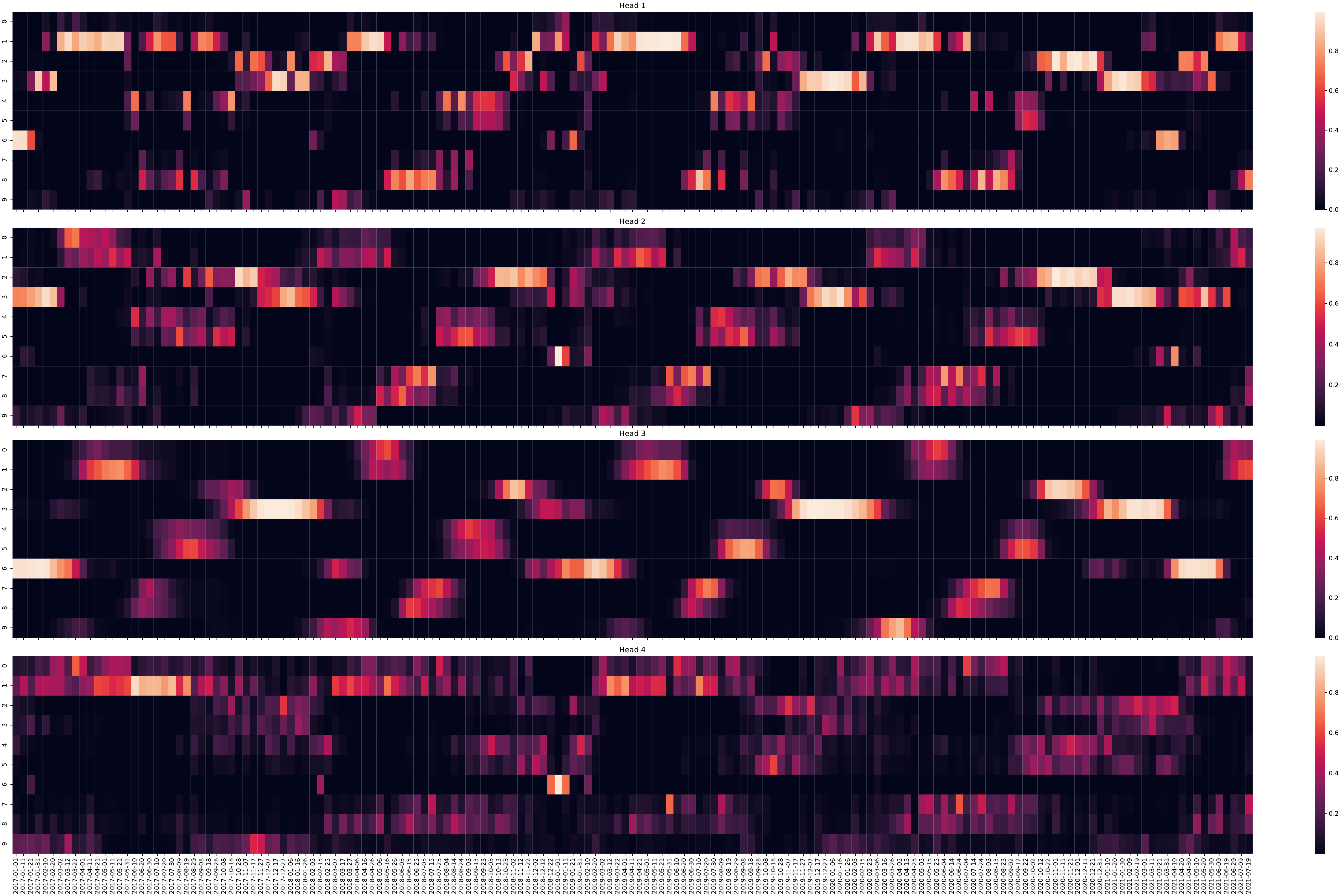}
\caption{\label{fig:attn-decod}Attention weights of the reconstruction decoder when reconstructing a time series from 2017-01-01 to 2021-07-19. Attention matrices plotted correspond to the average attention matrix obtained for each pixel of the SITS. Each matrix from top to bottom corresponds to a different head. On each score matrix, the column corresponds to the latent temporal feature, while the row corresponds to the dates to reconstruct.}
\end{figure*}

To gain a better understanding of the information encoded by ALISE, we conduct a qualitative analysis of the latent representations. For this purpose, we propose to study how latent information is used by the self-attention mechanism of the decoder during the  reconstruction process.  We note the pixel-level latent temporal vector, indexed by \(i\), as \(\mathbf{y_{(i,.)}}\).
To understand the importance of each latent temporal feature during the reconstruction process, the attention weights of each decoder head are displayed in \autoref{fig:attn-decod}. For improved visualization, the reconstruction decoder is asked to reconstruct in  \autoref{fig:attn-decod}, a regularly sampled time series from 2017-01-01 to 2021-07-19 with a step of 10 days. The reconstruction temporal grid considered here corresponds to the years observed during pre-training (2017-2020) as well as years outside the temporal extent of the pre-training data-set (2021).
In an attention matrix, a high attention score at a given row (indexed by \(i\)) and column (corresponding to a date \(d_j\)) indicates the importance of latent temporal feature \(\mathbf{y_{(i,.)}}\) for the reconstruction of the date \(d_j\). For each latent temporal feature (row), high attention scores (bright color) are often observed on specific narrow intervals, while attention weights are low outside of them. A notable finding is that for a latent temporal feature, these intervals are often separated by approximately one year. Although no annual periodicity is explicitly given as input to the decoder, dates to reconstruct spaced of 365 days  exhibit similar high attention score on the same latent temporal feature.
Furthermore, it is worth noting these annual periodic patterns are also observed for reconstructed dates not included in the pre-training (year 2021), suggesting that the model might have forecasting (extrapolation) abilities.

\begin{figure*}[htb]
\centering
\includegraphics[width=.9\linewidth]{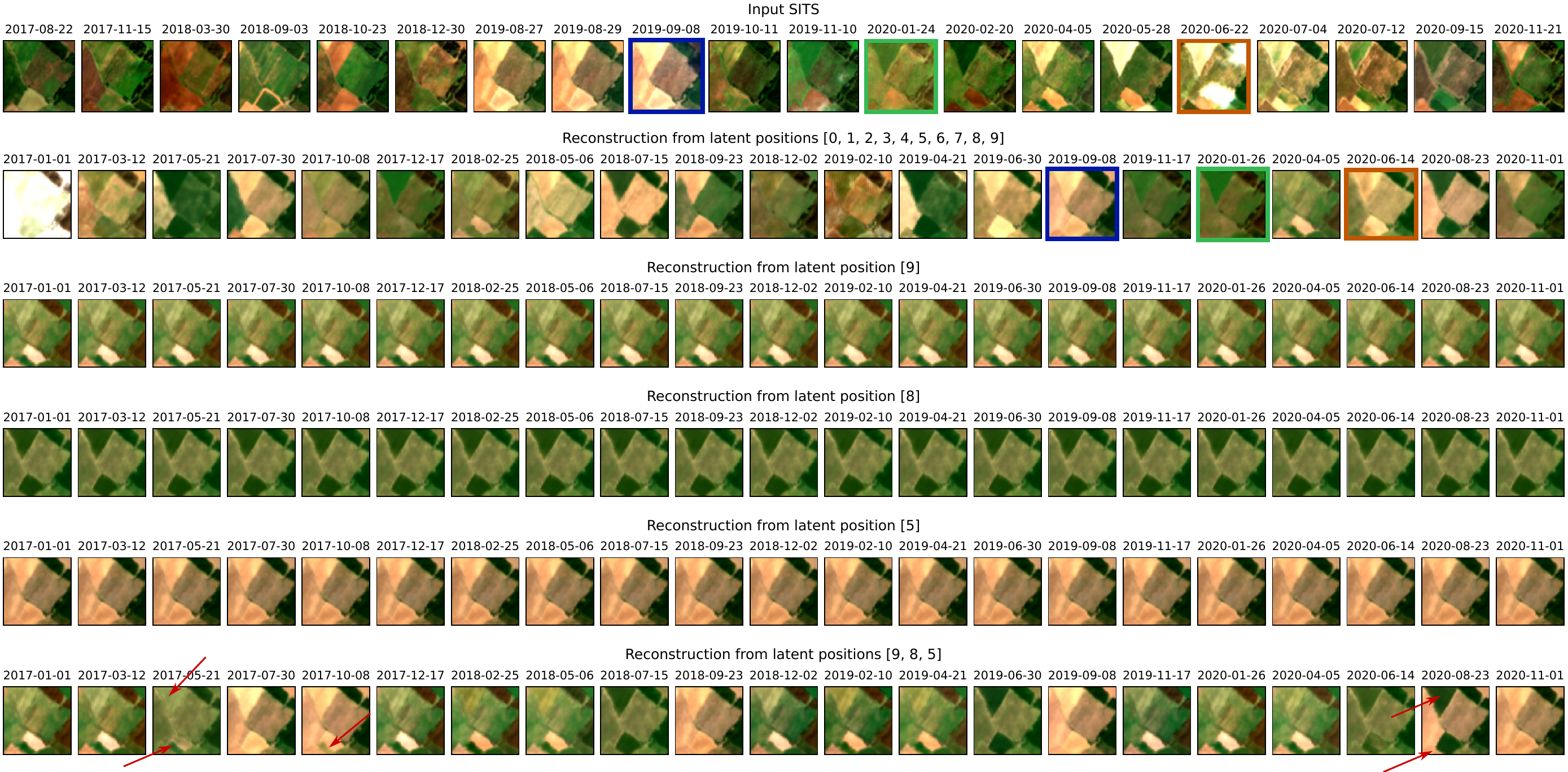}
\caption{\label{fig:rec-queries}Reconstructions obtained from the decoder given different latent temporal features configurations. First row: random dates from the input SITS. Then reconstruction obtained by using: all latent temporal features (second row), latent temporal feature n°9 (third row), latent temporal feature n°8 (fourth row), latent temporal feature n°5 (fifth row), latent temporal features 9,8 and 5 (bottom row).}
\end{figure*}

The influence of the latent temporal features on the reconstruction is also illustrated in \autoref{fig:rec-queries}. Specifically, the reconstruction is conducted with either all latent temporal features, a single latent temporal feature, or a triplet of features. The top row depicts random acquisition dates within the input SITS, while the next rows display the reconstruction of the decoder of the SITS on dates ranging from 2017-01-01 to 2021-11-01, with acquisitions spaced by 70-day intervals. Consequently, the image acquisitions of the first row are not temporally aligned with the rows below. Nevertheless, the predictions generated by the model using all latent temporal features (second row) and the input SITS (first row) are often consistent. For example, the acquisitions marked by the blue, green, and orange rectangles are temporally close to some of the input predictions. These latter acquisitions show coherent reconstructions, which highlights that the use of the temporal alignment projector does not lead to a significant loss of information. The following three rows in \autoref{fig:rec-queries} illustrate reconstruction results obtained when a single latent temporal feature is used during the decoding stage, \(\mathbf{y_{(9,.)}}\), \(\mathbf{y_{(8,.)}}\) and \(\mathbf{y_{(5,.)}}\), respectively. For each row, it can be observed that the model invariably generates the same image, revealing that the temporal dynamics are not reconstructed. Conversely, the temporal dynamic can be observed in the reconstructions obtained from only three latent temporal features (last row). For this case, the contribution of each latent temporal feature to the image reconstruction process is easily recognizable. In the proposed example, the reconstruction obtained from \(\mathbf{y_{(9,.)}}\) highly contributes to the reconstruction of winter images shown on the last row. The latent temporal features \(\mathbf{y_{(8,.)}}\) and \(\mathbf{y_{(5,.)}}\) seem to intervene for May to July and July to November, respectively. In this last row, the majority of the images appear to be highly similar to the reconstruction obtained from one single latent temporal feature. Nevertheless, a few acquisitions seem to be the obtained through the mix of information from different latent temporal features. For instance for 2017-05-21, it seems that the latent temporal features \(\mathbf{y_{(8,.)}}\) and \(\mathbf{y_{(9,.)}}\) are merged. Areas showing this combination are marked by the red arrows in \autoref{fig:rec-queries}.  From these observations we may consider the aligned representations as a novel basis for input SITS representation. From this point of view, latent temporal feature data would serve as prototype and the attention weights in the decoder would fulfill the role of membership degree.

\section{Conclusion}
\label{sec:conclusion}
This article discusses the notable challenges involved in learning to represent satellite image time series (SITS). In particular, our work paves the way for the construction of a FM for land surface monitoring using Earth observation.

This paper proposes a new SITS encoder named ALISE, which exploits spatial, spectral and temporal dimensions and generates aligned and fixed-size representations of irregular and unaligned multi-year SITS. ALISE is pre-trained using a new multi-view hybrid SSL pre-training task that combines MAE loss with instance discrimination losses. In addition, ALISE pre-training data-set (MMDC-EU) is a custom-built large-scale multi-year unlabeled dataset. The quality of ALISE's representation has been assessed on three downstream data-sets: PASTIS (crop segmentation), MultiSenGE (dense land cover segmentation) and the novel CropRot (crop change detection). Our results demonstrate the significant progress made with regard to the three representation characteristics considered: \textbf{easy to use}, \textbf{informative} and \textbf{generic}.

Firstly, ALISE provides aligned, fixed-size representations that preserve the spatial resolution of the input data. Consequently, ALISE representations can be easily exploited by a shallow classifier. In this paper, a single fully connected layer was used to perform two segmentation tasks (crop mapping and dense land cover).  Results have also shown that pre-trained and frozen ALISE outperforms the fully supervised approach when labeled data are scarce. The remarkable performance of frozen and pre-trained ALISE indicates that an important step has been taken towards the creation of \textbf{ready-to-use} SITS representations.
The production of aligned, fixed-size representations has been achieved through the use of a learnable query-based cross-attention mechanism. We have also provided the first qualitative study of the aligned latent temporal features obtained through this latter mechanism. It appears that each latent pseudo-date summarizes a specific part of the input SITS. To reconstruct a SITS, the pre-training decoder successfully recovers the annual periodicity of the SITS, whereas our temporal encoding does not rely on the day of the year.

Secondly, the quality of ALISE representations was compared with existing competitive works. Results obtained by pre-trained and frozen ALISE outperform Presto \cite{Tseng2023LightweightPT} and U-BARN \cite{dumeur-2024-self-super} for both semantic segmentation tasks. As a result, ALISE's representations may be considered more \textbf{informative} than other existing works. In addition, we have studied in depth our proposed hybrid SSL approach. Notably, the impact of the view generation method and the contribution of each loss has also been investigated. Our results show that most of the pre-training is driven by the cross-reconstruction task. Nevertheless, depending on how the view generation is performed, which is strongly influenced by \(t_w\), the invariance loss may or may not improve performances. This leads us to think that other view generation protocols could be investigated. Besides our experiments did not reveal a significant contribution from the covariance loss. These unexpected findings also highlight the important challenges that remain in applying ideas from the wider computer vision field to the specificities of SITS (temporal dynamics, physics of the measure, etc.). 

Thirdly, the \textbf{genericity} of the representations was assessed on three proposed downstream tasks. In addition to the great performances obtained in both segmentation tasks, our results demonstrate that the proposed aligned SITS representations can be used for downstream unsupervised change detection tasks. We also consider the proposed novel crop change detection data-set named \emph{CropRot}, as an important contribution to assess future FM on SITS. Besides, to learn \textbf{generic} representations, ALISE was pre-trained on a new large-scale and multi-year data-set. Nevertheless, the geographical diversity of the pre-training and downstream data-sets could be improved, since pre-training and downstream data only includes European geographical areas. The development of a scalable method, trained and evaluated on numerous geographical configurations, remains unexplored here.

It should also be noted, however, that this article does not address certain aspects.  For instance, ALISE memory consumption is quadratic with the temporal size of the input SITS. Therefore, lightweight architectures based on learnable queries \cite{yang2023gpvit,DBLP:conf/bmvc/CaiBNS23,pmlr-v139-jaegle21a} could be considered. Additionally, the construction of a global pre-training data-set as well as the study of the incorporation of thermal encoding \cite{Nyborg_2022_CVPR} to improve spatial scalability are worthy of interest. Lastly, a major remaining challenge in developing FM is the processing of multi-sensor data. For example, combining S1 data with S2 data is beneficial when optical data are unavailable due to unsuitable weather conditions. Furthermore, using different modalities in a multi-view SSL protocol is promising, and we might observe a greater contribution from instance discrimination losses in this context.

 \bibliographystyle{IEEEtran}
 {\scriptsize
\bibliography{local.bib}}

\appendix
\crefname{section}{Appendix}{Appendices}
\section*{Acknowledgments}
\label{sec:orge8b6aea}
This work was funded by the ANR-JCJC DeepChange project under Grant Agreement 20-CE23-0003 and by the EU Horizon Europe EvoLand project under grant agreement 101082130.
This work was partly performed using HPC resources from GENCI-IDRIS (Grant 2023-AD011014912)
This work was partly performed using HPC resources from CNES Computing Center.

\section{Appendix}
\label{sec:orgf1b58b7}
\subsection{CropRot additional information}
\label{app:rotcrop-confmat}
\begin{figure}[htbp]
\centering
\includegraphics[width=0.8\linewidth]{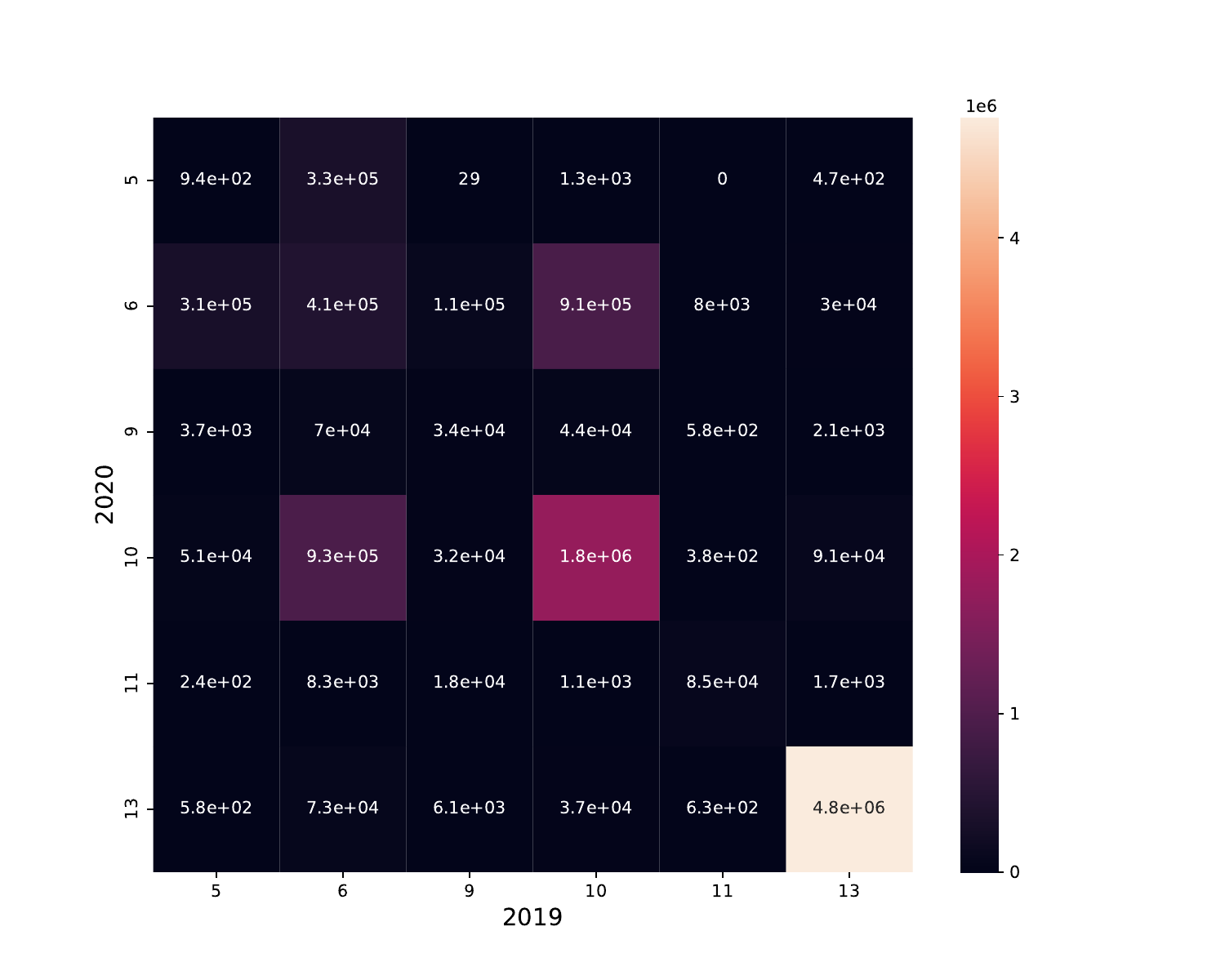}
\caption{\label{fig:confmat}Change matrix between years 2019 and 2020 on the crop classes. Classes correspondence is \{5: rapeseed, 6: cereal, 9 : sunflower, 10 : corn, 11 : rice, 13: grassland\}}
\end{figure}

The change matrix between 2019 and 2020 is represented by \Cref{fig:confmat}. As expected, the rate of change depends on the considered class. We observe important rotations between cereal and corn, while grassland mostly remains unchanged.
\subsection{Comparison with VicRegL}
\label{app:vicregl}
The proposed discriminative losses are similar to those of VicRegL \cite{bardes2022vicregl}. However, three notable modifications have been introduced. Firstly, unlike VicRegL, the invariance loss does not require any matching functions to realign the  pixels of both views since geometric augmentation is not performed. In our case, each embedded vector at the pixel level is compared with the embedded vector of the other view at the same spatial position. Secondly, the large SITS size strongly constrains the batch size, which differs from the larger batch values of VicRegL. In VicRegL, the covariance loss is computed for each pixel of the latent representation using the \(b\) samples of the batch. The final local covariance loss is the sum over the spatial dimensions \(h \times w\) of the pixel-level losses. Instead of estimating a covariance for each pixel, our covariance loss is estimated for the \(d_{emb}\) variables using \(b \times h \times w\) samples. Thirdly, the variance loss is not considered in our approach. If the variance was estimated by considering \(b \times h \times w\) samples, keeping the variance of each variable above a threshold would enforce a strong variability between pixels that might come from the same image. This loss could then deteriorate the spatial consistency of the representation.
\subsection{ALISE architecture}
\label{app:alise-archi}
\subsubsection{Other architecture hyper-parameters}
\label{sec:orgc65359a}
\begin{enumerate}
\item U-BARN
\label{sec:org984397e}
\Cref{tab:unet} and \Cref{tab:transfo-hp} describe the architectural hyper-parameters of the spatio-spectro-temporal encoder. 
\begin{table}[htbp]
\caption{\label{tab:unet}Hyper-parameters of the architecture of the Unet encoder, with B and T respectively the batch and temporal dimensions. The \emph{down block} architecture is detailed in \cite{dumeur-2024-self-super}}
\centering
\begin{tabular}{|l|l|l|}
\hline
Block Name & Input dimensions & Output dimensions\\
\hline
Input Convolution & (B*T,64,64,10) & (B*T,64,64,64)\\
Down Block 1 & (B*T,64,64,64) & (B*T,32,32,64)\\
Down Block 2 & (B*T,32,32,64) & (B*T,16,16,64)\\
Down Block 3 & (B*T,16,16,64) & (B*T,8,8,128)\\
\hline
\end{tabular}
\end{table}
\begin{table}[htbp]
\caption{\label{tab:transfo-hp}Architectural hyper-parameters of the Transformer in U-BARN}
\centering
\begin{tabular}{|p{0.10\linewidth}|p{0.15\linewidth}|p{0.15\linewidth} |p{0.15\linewidth}|p{0.15\linewidth}|p{0.15\linewidth}|}
\hline
N\textsubscript{layers} & N\textsubscript{head} & attn\textsubscript{dropout} & dropout & d\textsubscript{model} & d\textsubscript{hidden}\\
\hline
3 & 4 & 0.1 & 0.1 & 64 & 128\\
\hline
\end{tabular}
\end{table}

\item Temporal projector.
\label{sec:org7c35ab3}

The temporal projector is composed of a lightweight multi-head cross-attention mechanism with two heads. Inspired by the attention mechanism proposed in \cite{garnot20_light_tempor_self_class_satel}, the channels of the input embeddings are distributed among the heads.
\end{enumerate}

\subsection{MultiSenGE data distribution}
\label{app:msenge-classes}
\begin{center}
\includegraphics[width=\linewidth]{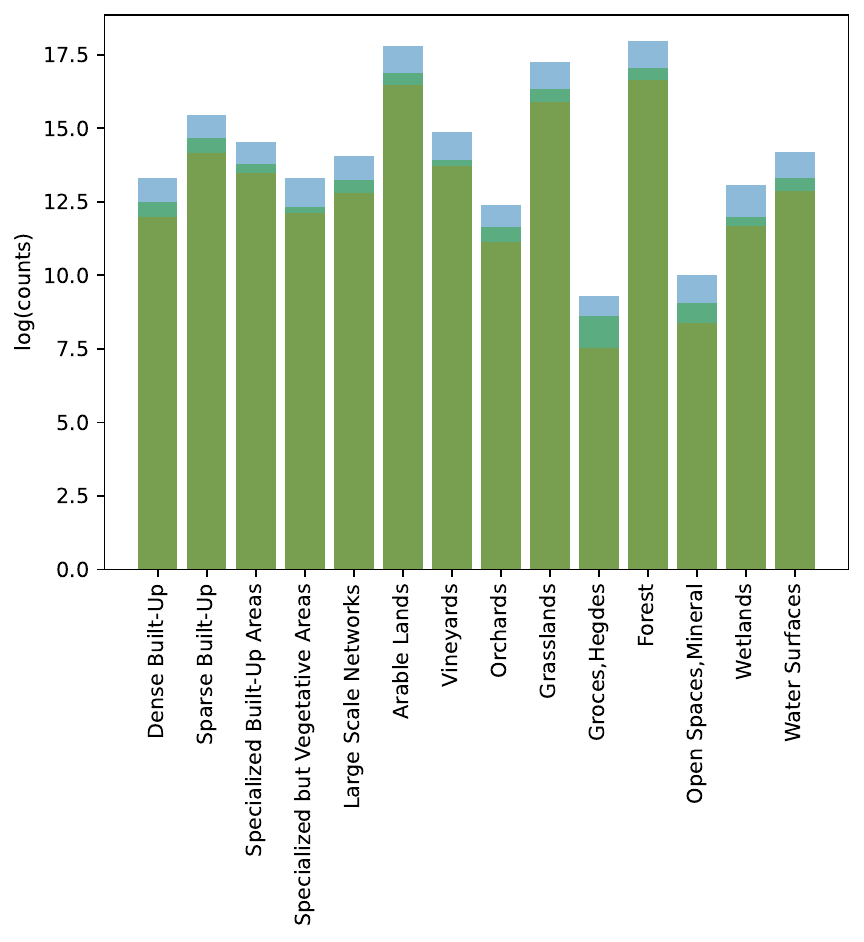}
\end{center}
\end{document}